\documentclass[preprints,article,accept,moreauthors,pdftex]{Definitions/mdpi} 

\firstpage{1} 
\pubvolume{xx}
\issuenum{1}
\articlenumber{5}
\pubyear{2019}
\copyrightyear{2019}
\history{Received: date; Accepted: date; Published: date}

\usepackage{subcaption}
\usepackage{colortbl}
\usepackage{pgfplots}
\usepackage{adjustbox}
\usepackage{longtable}
\pgfplotsset{compat=1.9}
\DeclareMathOperator{\E}{\mathbb{E}}
\Title{Small-Object Detection in Remote Sensing Images with End-to-End Edge-Enhanced GAN and Object Detector Network}

\Author{Jakaria Rabbi $^{1,}$*\orcidA{}, Nilanjan Ray $^{1}$\orcidC{}, Matthias Schubert $^{2}$, Subir Chowdhury $^{3}$\orcidB{} and Dennis Chao $^{3}$}

\AuthorNames{Jakaria Rabbi, Nilanjan Ray, Matthias Schubert, Subir Chowdhury and Dennis Chao}

\address{%
$^{1}$ \quad Department of Computing Science,
2-32 Athabasca Hall,
University of Alberta, 
Edmonton, AB T6G 2E8, Canada; nray1@ualberta.ca\\
$^{2}$ \quad Institute for Informatic, Ludwig-Maximilians-Universität München, Oettingenstraße 67,
D-80333 Munich, Germany; schubert@dbs.ifi.lmu.de\\
$^{3}$ \quad Alberta Geological Survey, Alberta Energy Regulator, Edmonton, Alberta, Canada; subir.chowdhury@aer.ca; dennis.chao@aer.ca}

\corres{Correspondence: jakaria@ualberta.ca}

\abstract{The detection performance of small objects in remote sensing images has not been satisfactory compared to large objects, especially in low-resolution and noisy images. A generative adversarial network (GAN)-based model called enhanced super-resolution GAN (ESRGAN) showed remarkable image enhancement performance, but reconstructed images usually miss high-frequency edge information. Therefore, object detection performance showed degradation for small objects on recovered noisy and low-resolution remote sensing images. Inspired by the success of edge enhanced GAN (EEGAN) and ESRGAN, we applied a new edge-enhanced super-resolution GAN (EESRGAN) to improve the quality of remote sensing images and used different detector networks in an end-to-end manner where detector loss was backpropagated into the EESRGAN to improve the detection performance. We proposed an architecture with three components: ESRGAN, EEN, and Detection network. We used residual-in-residual dense blocks (RRDB) for both the ESRGAN and EEN, and for the detector network, we used a faster region-based convolutional network (FRCNN) (two-stage detector) and a single-shot multibox detector (SSD) (one stage detector). Extensive experiments on a public (car overhead with context) dataset and another self-assembled (oil and gas storage tank) satellite dataset showed superior performance of our method compared to the standalone state-of-the-art object detectors.}


\begin{document}


\abbreviations{The following acronyms are used in this paper:\\

\noindent 
\begin{longtable}{@{}ll}
SRCNN & Single image Super-Resolution Convolutional  Neural  Network\\
VDSR & Very Deep Convolutional Networks \\
GAN & Generative Adversarial Network \\
SRGAN & Super-Resolution Generative Adversarial Network \\
ESRGAN & Enhanced Super-Resolution Generative Adversarial Network \\
EEGAN & Edge-Enhanced Generative Adversarial Network \\
EESRGAN & Edge-Enhanced Super-Resolution Generative Adversarial Network \\
RRDB & Residual-in-Residual Dense Blocks \\
EEN & Edge-Enhancement Network \\
SSD & Single-Shot MultiBox Detector \\
YOLO & You Only Look Once  \\
CNN & Convolutional  Neural  Network\\
R-CNN & Region-based Convolutional  Neural  Network \\ 
FRCNN & Faster Region-based Convolutional  Neural  Network \\ 
VGG & Visual Geometry Group \\
BN & Batch Normalization \\
MSCOCO & Microsoft Common Objects in Context\\
OGST &  Oil and Gas Storage Tank \\
COWC & Car Overhead With Context \\
GSD & Ground Sampling Distance \\
G & Generator \\
D & Discriminator \\
ISR &  Intermediate Super-Resolution \\   
SR & Super-Resolution \\ 
HR & High-Resolution \\
LR & Low-Resoluton \\
GT & Ground Truth \\
FPN & Feature Pyramid Network \\
RPN & Region Proposal Network \\ 
AER & Alberta Energy Regulator \\
AGS & Alberta Geological Survey \\
AP & Average Precision \\
IoU & Intersection over Union \\
TP & True Positive \\
FP & False Positive \\
FN & False Negative \\

\end{longtable}}

\section{Introduction}
\subsection{Problem Description and Motivation}
Object detection on remote sensing imagery has numerous prospects in various fields, such as environmental regulation, surveillance, military \cite{military1,military2}, national security, traffic, forestry \cite{forestrySeedling}, oil and gas activity monitoring. 
There are many methods for detecting and locating objects from images, which are captured using satellites or drones. However, detection performance is not satisfactory for noisy and low-resolution (LR) images, especially when the objects are small \cite{TinyObjectFilter}. Even on high-resolution (HR) images, the detection performance for small objects is lower than that for large objects \cite{small_object_enhance}. \par
Current state-of-the-art detectors have excellent accuracy on benchmark datasets, such as ImageNet \cite{imagenet} and Microsoft common objects in context (MSCOCO) \cite{mscoco}. These datasets consist of everyday natural images with distinguishable features and comparatively large objects.\par
On the other hand, there are various objects in satellite images like vehicles, small houses, small oil and gas storage tanks etc., only covering a small area \cite{TinyObjectFilter}. The state-of-the-art detectors \cite{FRCNN,RetinaNet,SSD,yolo} show a significant performance gap between LR images and their HR counterparts due to a lack of input features for small objects \cite{vehicleSuperResolution}. In addition to the general object detectors, researchers have proposed specialized methods, algorithms, and network architectures to detect particular types of objects from satellite images such as vehicles \cite{Vehicle1,vehicle2}, buildings \cite{building}, and storage tanks \cite{storage_tank}. These methods are object-specific and use fixed resolution for feature extraction and detection. \par
To improve detection accuracy on remote sensing images, researchers have used deep convolutional neural network (CNN)-based super-resolution (SR) techniques to generate artificial images and then detect objects \cite{small_object_enhance,vehicleSuperResolution}. Deep CNN-based SR techniques such as single image super-resolution convolutional networks (SRCNN) \cite{SRCNN} and accurate image super-resolution using very deep convolutional networks (VDSR) \cite{VDSR} showed excellent results on generating realistic HR imagery from LR input data. Generative Adversarial Network (GAN)-based \cite{GAN} methods such as super-resolution GAN (SRGAN) \cite{SRGAN} and enhanced super-resolution GAN (ESRGAN) \cite{ESRGAN} showed remarkable performance in enhancing LR images with and without noise. These models have two subnetworks: a generator and a discriminator. Both  subnetworks consist of deep CNNs. Datasets containing HR and LR image pairs are used for training and testing the models. The generator generates HR images from LR input images, and the discriminator predicts whether generated image is a real HR image or an upscaled LR image. After sufficient training, the generator generates HR images that are similar to the ground truth HR images, and the discriminator cannot correctly discriminate between real and fake images anymore. \par
Although the resulting images look realistic, the compensated high-frequency details such as image edges may cause inconsistency with the HR ground truth images \cite{EEGAN}. Some works showed that this issue negatively impacts land cover classification results \cite{edge_land_cover1,edge_land_cover2}. Edge information is an important feature for object detection \cite{edge_important}, and therefore, this information needs to be preserved in the enhanced images for acceptable detection accuracy. \par
In order to obtain clear and distinguishable edge information, researchers proposed several methods using separate deep CNN edge extractors \cite{edge_CNN1, edge_CNN2}. The results of these methods are sufficient for natural images, but the performance degrades on LR and noisy remote sensing images \cite{EEGAN}. A recent method \cite{EEGAN} used the GAN-based edge-enhancement network (EEGAN) to generate a visually pleasing result with sufficient edge information. EEGAN employs two subnetworks for the generator. One network generates intermediate HR images, and the other network generates sharp and noise-free edges from the intermediate images. The method uses a Laplacian operator \cite{laplacian_operator} to extract edge information and in addition, it uses a mask branch to obtain noise-free edges. This approach preserves sufficient edge information, but sometimes the final output images are blurry compared to a current state-of-the-art GAN-based SR method \cite{ESRGAN} due to the noises introduced in the enhanced edges that might hurt object detection performance. \par
Another important issue with small-object detection is the huge cost of HR imagery for large areas. Many organizations are using very high-resolution satellite imagery to fulfill their purposes. When it comes to continuous monitoring of a large area for regulation or traffic purposes, it is costly to buy HR imagery frequently.
Publicly available satellite imagery such as Landsat-8 \cite{landsat-8} (30 m/pixel) and Sentinel-2 \cite{sentinel-2} (10 m/pixel) are not suitable for detecting small objects due to the high ground sampling distance (GSD). Detection of small objects (e.g., oil and gas storage tanks and buildings) is possible from commercial satellite imagery such as 1.5-m GSD SPOT-6 imagery but the detection accuracy is low compared to HR imagery, e.g., 30-cm GSD DigitalGlobe imagery in Bing map.
\par
We have identified two main problems to detect small-objects from satellite imagery.  First, the accuracy of small-object detection is lower compared to large objects, even in HR imagery due to sensor noise, atmospheric effects, and geometric distortion. Secondly, we need to have access to HR imagery, which is very costly for a vast region with frequent updates. Therefore, we need a solution to increase the accuracy of the detection of smaller objects from LR imagery. To the best of our knowledge, no work employed both SR network with edge-enhancement and object detector network in an end-to-end manner, i.e, using joint optimization to detect small remote sensing objects.

\begin{figure}[H]
\centering
\captionsetup{width=.9\linewidth}
\renewcommand{\arraystretch}{0.01}
\centering
\begin{tabular}{@{} c @{} | @{} c @{} | @{} c @{} | @{} c @{} }
\rotatebox{90}{\hspace{23pt}\framebox{COWC Dataset}} \hspace{8pt} &
\includegraphics[width = 1.5in]{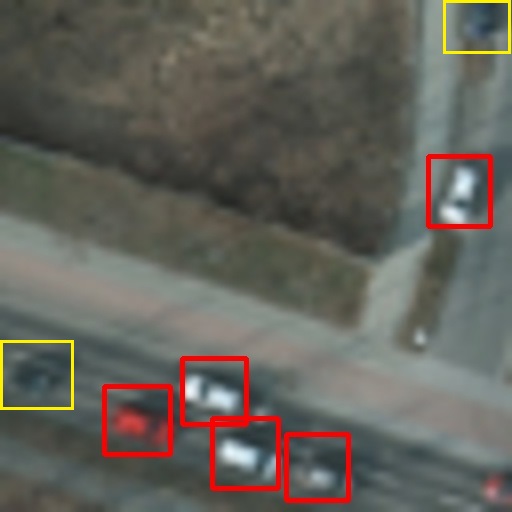} &
\includegraphics[width = 1.5in]{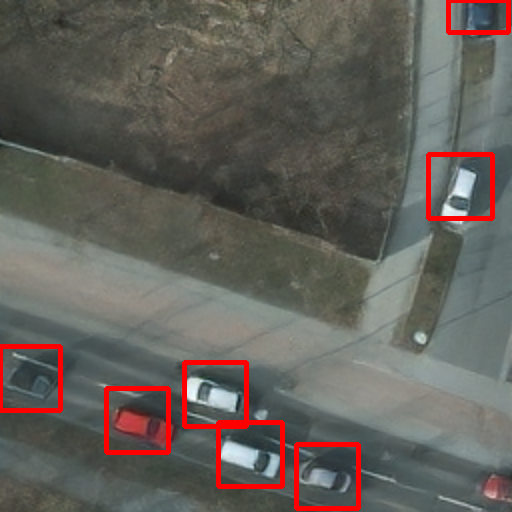} &
\begin{tikzpicture}
\begin{axis}[
    width=2.6in,
    height=1.6in,
    xlabel={IoU},
    ylabel={AP},
    xmin=0.5, xmax=1,
    ymin=0, ymax=100,
    xtick={0.50,0.60,0.70,0.80,0.90, 1.0},
    yticklabel={$\pgfmathprintnumber{\tick}\%$},
    legend pos=north west,
    legend cell align=left,
    grid=both,
    grid style={line width=.1pt, draw=gray!20},
    legend style={legend pos=south west, draw=none, fill=none, font=\fontsize{7}{5}\selectfont}
]

\addplot[
    color=red
    ]
    coordinates {
    (0.50,96.2)(.55,96.2)(0.60,96.1)(0.65,93.39)(0.70,89.92)(0.75,76.9)(0.80,50.99)(0.85,30.19)(0.90,18.93)(0.95,4.36)(1.0,0)
    };
    \addlegendentry{FRCNN (LR)}
    
\addplot[
    color=blue
    ]
    coordinates {
    (0.50,98.8)(.55,98.4)(0.60,98.13)(0.65,98.07)(0.70,97.89)(0.75,97.81)(0.80,97.3)(0.85,94.13)(0.90,88.13)(0.95,67.6)(1.0,0)
    };
    \addlegendentry{EESRGAN-FRCNN-End-to-End (LR)}
    
\end{axis}
\end{tikzpicture} \\
\rotatebox{90}{\hspace{24pt}\framebox{OGST Dataset}} \hspace{8pt} &
\includegraphics[width = 1.5in]{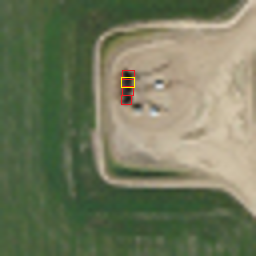} &
\includegraphics[width = 1.5in]{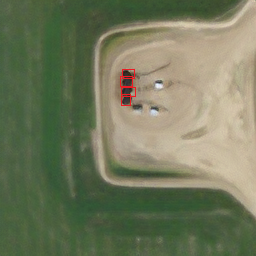} &
\begin{tikzpicture}
\begin{axis}[
    width=2.6in,
    height=1.6in,
    xlabel={IoU},
    ylabel={AP},
    xmin=0.5, xmax=1,
    ymin=0, ymax=100,
    xtick={0.50,0.60,0.70,0.80,0.90, 1.0},
    yticklabel={$\pgfmathprintnumber{\tick}\%$},
    legend pos=north west,
    legend cell align=left,
    grid=both,
    grid style={line width=.1pt, draw=gray!20},
    legend style={legend pos=south west, draw=none, fill=none, font=\fontsize{7}{5}\selectfont}
]

\addplot[
    color=red
    ]
    coordinates {
    (0.50,93.32)(.55,93.32)(0.60,93.32)(0.65,93.32)(0.70,92.54)(0.75,88.03)(0.80,81.4)(0.85,71.07)(0.90,50.89)(0.95,14.05)(1.0,0)
    };
    \addlegendentry{FRCNN (LR)}
    
\addplot[
    color=blue
    ]
    coordinates {
    (0.50,97.1)(.55,97.1)(0.60,96.56)(0.65,96.56)(0.70,96.56)(0.75,93.14)(0.80,89.89)(0.85,81.59)(0.90,58.64)(0.95,25.54)(1.0,0)
    };
    \addlegendentry{EESRGAN-FRCNN-End-to-End (LR)}
    
\end{axis}
\end{tikzpicture}
 \\
\arrayrulecolor{white}
\hline
\hline
\hline
\multicolumn{1}{c}{ } & \multicolumn{1}{c}{(I) LR image} & \multicolumn{1}{c}{(II) SR image} & \multicolumn{1}{c}{\hspace{1cm}(III) AP vs IoU curves}
\end{tabular}
 \vspace{\floatsep}
 \\
\centering
\caption{Detection on LR (low-resolution) images (60cm/pixel) is shown in (I); in (II), we show the detection on generated SR (super-resolution) images (15 cm/pixel). The first row of this figure represents the COWC (car overhead with context) dataset \cite{cowc}, and the second row represents the OGST (oil and gas storage tank) dataset \cite{oil-gas-tank-dataset}. AP (average precision) values versus different IoU (intersection over union) values for the LR test set and generated SR images from the LR images are shown in (III) for both the datasets. We use FRCNN (faster region-based CNN) detector on LR images for detection. Then instead of using LR images directly, we use our proposed end-to-end EESRGAN (edge-enhanced SRGAN) and FRCNN architecture (EESRGAN-FRCNN) to generate SR images and simultaneously detect objects from the SR images. The red bounding boxes represent true positives, and yellow bounding boxes represent false negatives. IoU=0.75 is used for detection.}
\label{fig:AP_IoU_image}
\end{figure}
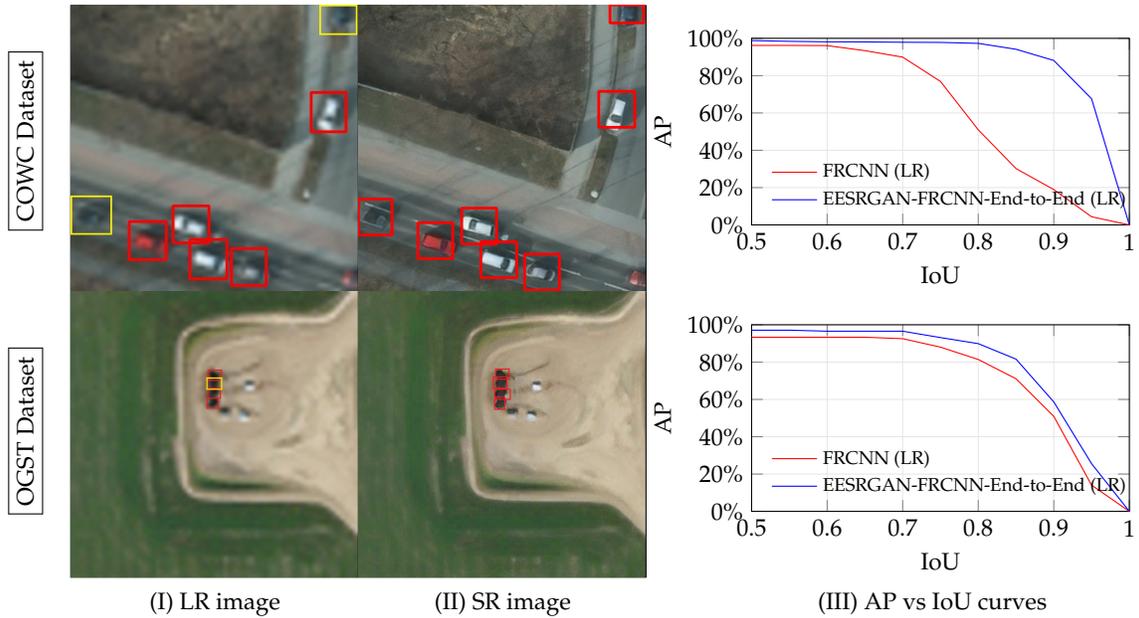

In this paper, we propose an end-to-end architecture where object detection and super-resolution is performed simultaneously. Figure \ref{fig:AP_IoU_image} shows the significance of our method. State-of-the-art detectors miss objects when trained on the LR images; in comparison, our method can detect those objects. The detection performance improves when we use SR images for the detection of objects from two different datasets. Average precision (AP) versus different intersection over union (IoU) values (for both LR and SR) are plotted to visualize overall performance on test datasets. From figure \ref{fig:AP_IoU_image}, we observe that for both the datasets, our proposed end-to-end method yields significantly better IoU values for the same AP. In section \ref{ssec:Eval_mat}, we discuss AP and IoU in more detail and these results are discussed in section \ref{sec:results}.

\subsection{Contributions of Our Method}
Our proposed architecture consists of two parts: EESRGAN network and a detector network. Our approach is inspired by EEGAN and ESRGAN networks and showed a remarkable improvement over EEGAN to generate visually pleasing SR satellite images with enough edge information. We employed a generator subnetwork, a discriminator subnetwork, and an edge-enhancement subnetwork \cite{EEGAN} for the SR network. For the generator and edge-enhancement network, we used residual-in-residual dense blocks (RRDB) \cite{ESRGAN}. These blocks contain multi-level residual networks with dense connections that showed good performance on image enhancement. \par 
We used a relativistic discriminator \cite{relativisticGAN} instead of a normal discriminator. Besides GAN loss and discriminator loss, we employed Charbonnier loss \cite{Charbonnier1994TwoDH} for the edge-enhancement network. Finally, we used different detectors \cite{FRCNN,SSD} to detect small objects from the SR images. The detectors acted like the discriminator as we backpropagated the detection loss into the SR network and, therefore, it improved the quality of the SR images. \par
We created the oil and gas storage tank (OGST) dataset \cite{oil-gas-tank-dataset} from satellite imagery (Bing map), which has 30 cm and 1.2 m GSD. The dataset contains labeled oil and gas storage tanks from the Canadian province of Alberta, and we detected the tanks on SR images. Detection and counting of the tanks are essential for the Alberta Energy Regulator (AER) \cite{AER} to ensure safe, efficient, orderly, and environmentally responsible development of energy resources. Therefore, there is a potential use of our method for detecting small objects from LR satellite imagery. The OGST dataset is available on Mendeley \cite{oil-gas-tank-dataset}. \par

In addition to the OGST dataset, we applied our method on the publicly available car overhead with context (COWC) \cite{cowc} dataset to compare the performance of detection for varying use-cases. During training, we used HR and LR image pairs but only required LR images for testing.
Our method outperformed standalone state-of-the-art detectors for both datasets. \par
The remainder of this paper is structured as follows. We discuss related work in section \hyperref[sec:relatedworks]{2}. In section \hyperref[sec:method]{3}, we introduce our proposed method and describe every part of the method. The description of datasets and experimental results are shown in section \hyperref[sec:results]{4}, final discussion is stated in section \hyperref[sec:discussion]{5} and section \hyperref[sec:conclusions]{6} concludes our paper with a summary.

\section{Related Works}
\label{sec:relatedworks}
Our work consists of an end-to-end edge enhanced image SR network with an object detector network.  In this section, we  discuss existing methods related to our work.
\subsection{Image Super-Resolution}
Many methods were proposed on SR using deep CNNs. Dong et al. proposed super-resolution CNN (SRCNN) \cite{SRCNN} to enhance LR images in an end-to-end training outperforming previous SR techniques. The deep CNNs for SR evolved rapidly, and researchers introduced residual blocks \cite{SRGAN}, densely connected networks \cite{dense_network}, and residual dense block \cite{residual_dense_network} for improving SR results. He et al. \cite{He_2015} and Lim et al. \cite{EDSR} used deep CNNs without the batch normalization (BN) layer and observed significant performance improvement and stable training with a deeper network. These works were done on natural images. \par
Liebel et al. \cite{super-resolution-multispectral} proposed deep CNN-based SR network for multi-spectral remote sensing imagery. Jiang et al. \cite{EEGAN} proposed a new SR architecture for satellite imagery that was based on GAN. They introduced an edge-enhancement subnetwork to acquire smooth edge details in the final SR images.

\subsection{Object Detection}
Deep learning-based object detectors can be categorized into two subgroups, region-based CNN (R-CNN) models that employ two-stage detection and uniform models using single stage detection \cite{detectors_two_type}. Two-stage detectors comprise of R-CNN \cite{RCNN}, Fast R-CNN \cite{Fast_RCNN}, Faster R-CNN \cite{FRCNN} and the most used single stage detectors are SSD \cite{SSD}, You only look once (YOLO) \cite{yolo} and RetinaNet \cite{RetinaNet}. In the first stage of a two-stage detector, regions of interest are determined by selective search or a region proposal network. Then, in the second stage, the selected regions are checked for particular types of objects and minimal bounding boxes for the detected objects are predicted. In contrast, single-stage detectors omit the region proposal network and run detection on a dense sampling of all possible locations. Therefore, single-stage detectors are faster but, usually less accurate. RetinaNet \cite{RetinaNet} uses a focal loss function to deal with the data imbalance problem caused by many background objects and often showed similar performance as the two-stage approaches.
\par
Many deep CNN-based object detectors were proposed on remote sensing imagery to detect and count small objects, such as vehicles \cite{Vehicle1, vehicle_related_work, vehicle_related_work1}. Tayara et al. \cite{Vehicle1} introduced a convolutional regression neural network to detect vehicles from satellite imagery. Furthermore, a deep CNN-based detector was proposed \cite{vehicle_related_work} to detect multi oriented vehicles from remote sensing imagery. A method combining a deep CNN for feature extraction and a support vector machine (SVM) for object classification was proposed \cite{vehicle_related_work1}. Ren et al. \cite{small-object} modified the faster R-CNN detector to detect small objects in remote sensing images. They changed the region proposal network and incorporated context information into the detector. Another modified faster R-CNN detector was proposed by Tang et al. \cite{vehicle-hard-negative}. They used a hyper region proposal network to improve recall and used a cascade boosted classifier to verify candidate regions. This classifier can reduce false detection by mining hard negative examples. \par
An SSD-based end-to-end airplane detector with transfer learning was proposed, where, the authors used a limited number of airplane images for training \cite{plane-ssd}. They also proposed a method to solve the input size restrictions by dividing a large image into smaller tiles. Then they detected objects on smaller tiles and finally, mapped each image tile to the original image. They showed that their method performed better than the SSD model. In \cite{yolo-parameter-tuning}, the authors showed that finding a suitable parameter setting helped to boost the object detection performance of convolutional neural networks on remote sensing imagery. They used YOLO \cite{yolo} as object detector to optimize the parameters and infer the results.
\par
In \cite{forestrySeedling}, the authors detected conifer seedlings along recovering seismic lines from drone imagery. They used a dataset from different seasons and used faster R-CNN to infer the detection accuracy. There is another work \cite{palm-tree-cnn} related to plant detection, where authors detected palm trees from satellite imagery using sliding window techniques and an optimized convolutional neural network. \par

Some works produced excellent results in detecting small objects. Lin et al. \cite{FPNresnet50} proposed feature pyramid networks, which is a top-down architecture with lateral connections. The architecture could build high-level semantic feature maps at all scales. These feature maps boosted the object detection performance, especially for small object detection, when used as a feature extractor for faster R-CNN. Inspired by the receptive fields in human visual systems, Liu et al. \cite{RFBNet} proposed a receptive field block (RFB) module that used the relationship between the size and eccentricity of receptive fields to enhance the feature discrimination and robustness.  Hence, the module increased the detection performance of objects with various sizes when used as the replacement of the top convolutional layers of SSD.  \par

A one-stage detector called single-shot refinement neural network (RefineDet) \cite{refineDet} was proposed to increase the detection accuracy and also enhance the inference speed. The detector worked well for small object detection. RefineDet used two modules in its architecture: an anchor refinement module to remove negative anchors and an object detection module that took refined anchors as the input. The refinement helped to detect small objects more efficiently than previous methods. In \cite{fssd}, feature fusion SSD (FSSD) was proposed where features from different layers with different scales were concatenated together, and then some downsampling blocks were used to generate new feature pyramids. Finally, the features were fed to multibox detector for prediction. The feature fusion in FSSD increased the detection performance for both large and small objects.
Zhu et al. \cite{scratchdet} trained single-shot object detectors from scratch and obtained state-of-the-art performance on various benchmark datasets. They removed the first downsampling layer of SSD and introduced root block (with modified convolutional filters) to exploit more local information from an image. Therefore, the detector was able to extract powerful features for small object detection. \par
All of the aforementioned works were proposed for natural images. A method related to small object detection on remote sensing imagery was proposed by Yang et al. \cite{small-ship}. They used modified faster R-CNN to detect both large and small objects. They proposed rotation dense feature pyramid networks (R-DFPN), and the use of this network helped to improve the detection performance of small objects. \par
There is an excellent review paper by Zhao et al. \cite{object-review}, where the authors showed a thorough review of object detectors and also showed the advantages and disadvantages of different object detectors. The effect of object size was also discussed in the paper. Another survey paper about object detection in remote sensing images by Li et al. \cite{object-review-remote} showed review and comparison of different methods.

\subsection{Super-resolution along with Object Detection}
The positive effects of SR on object detection tasks was discussed in \cite{small_object_enhance} where the authors used remote sensing datasets for their experiments. Simultaneous CNN-based image enhancement with object detection using single-shot multibox detector (SSD) \cite{SSD} was done in \cite{enhance-detect1}. Haris et al. \cite{enhance-detect2} proposed a GAN-based generator to generate a HR image from a LR image and then used a multi-task network as a discriminator and also for localization and classification of objects. These works were done on natural images, and LR and HR image pairs were required. In another work \cite{vehicleSuperResolution}, a method using simultaneous super-resolution with object detection on satellite imagery was proposed. The SR network in this approach was inspired by the cycle-consistent adversarial network \cite{cyclegan}. A modified faster R-CNN architecture was used to detect vehicles from enhanced images produced by the SR network.

\section{Method}
\label{sec:method}
In this paper, we aim to improve the detection performance of small objects on remote sensing imagery. Towards this goal, we propose an end-to-end network architecture that consists of two modules: A GAN based SR network and a detector network. The whole network is trained in an end-to-end manner and HR and LR image pairs are needed for training. \par
\begin{figure}[H]
\centering
\includegraphics[width=15.5cm, height=9.5cm]{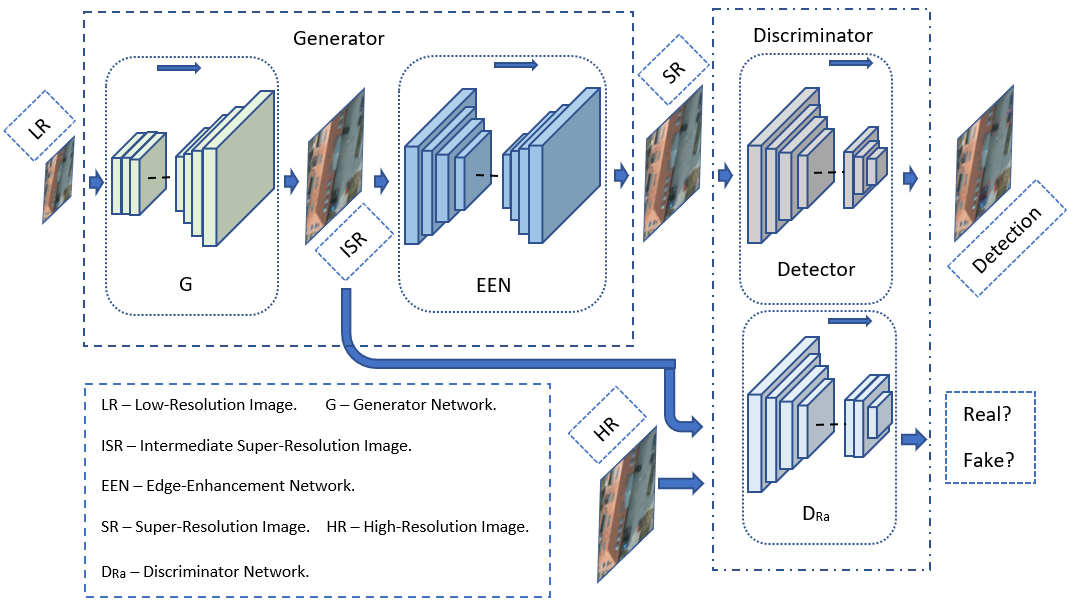}
\caption{Overall network architecture with a generator and a discriminator module.}
\label{fig:overallNetwork}
\end{figure}

 The SR network has three components: generator (G), discriminator ($D_{Ra}$), and edge-enhancement network (EEN). Our method uses end-to-end training as the gradient of the detection loss from the dectector is backpropagated into the generator. Therefore, the detector also works like a discriminator and encourages the generator G to generate realistic images similar to the ground truth. Our entire network structure can also be divided into two parts: A generator consisting of the EEN and a discriminator, which includes the $D_{Ra}$ and the detector network. In figure \ref{fig:overallNetwork}, we show the role of the detector as a discriminator. \par

The generator G generates intermediate super-resolution (ISR) images, and then final SR images are generated after applying the EEN network. The discriminator ($D_{Ra}$) discriminates between ground truth (GT) HR images and ISR. The inverted gradients of $D_{Ra}$ are backpropagated into the generator G in order to create SR images allowing for accurate object detection. Edge information is extracted from ISR, and the EEN network enhances these edges. Afterwards, the enhanced edges are again added to the ISR after subtracting the original edges extracted by the Laplacian operator and we get the output SR images with enhanced edges. Finally, we detect objects from the SR images using the detector network. \par

We use two different loss functions for EEN: one compares the difference between SR and ground truth images, and the other compares the difference between the extracted edge from ISR and ground truth. We also use the VGG19 \cite{vgg19} network for feature extraction that is used for perceptual loss \cite{ESRGAN}. Hence, it generates more realistic images with more accurate edge information. We divide the whole pipeline as a generator, and a discriminator, and these two components are elaborated in the following.

\subsection{Generator}
Our generator consists of a generator network G and an edge-enhancement network EEN. In this section, we describe the architectures of both  networks and the corresponding loss function.
\subsubsection{Generator Network $G$}

\begin{figure}[H]
\centering
\includegraphics[width=14 cm, height=5cm]{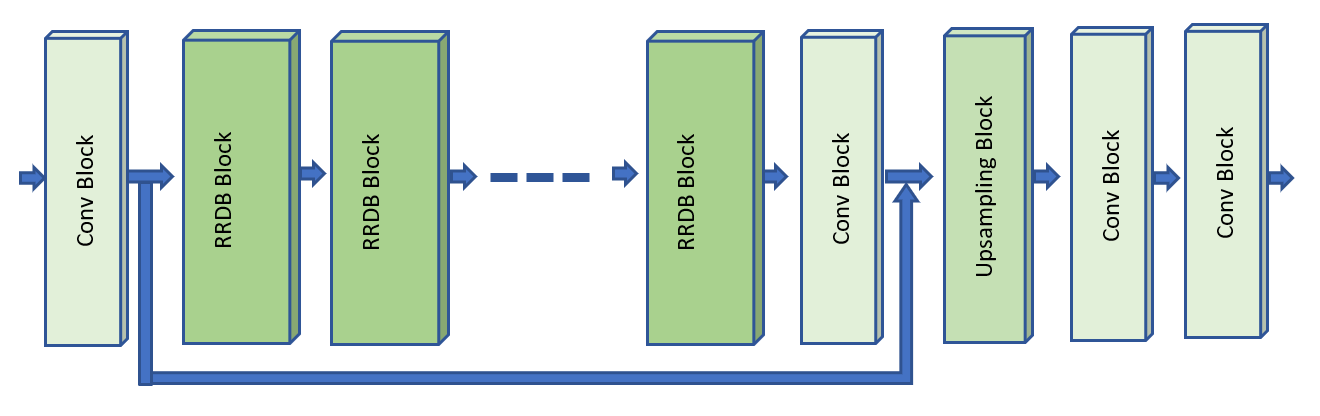}
\caption{Generator $G$ with RRDB (residual-in-residual dense blocks), convolutional and upsampling blocks.}
\label{fig:generator}
\end{figure}

\begin{figure}[H]
\centering
\begin{subfigure}[b]{0.6\textwidth}
\includegraphics[width=1\linewidth, height=5cm]{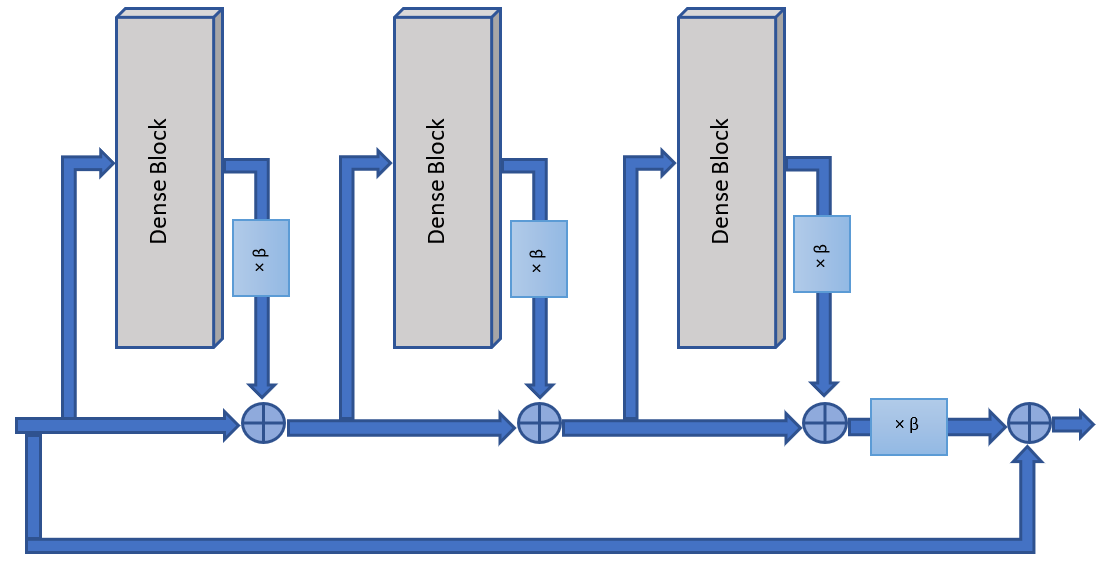}
\caption{RRDB from generator.} 
\label{fig:RRDB}
\vspace*{6mm}
\end{subfigure}

\begin{subfigure}[b]{0.7\textwidth}
\includegraphics[width=1\linewidth, height=5cm]{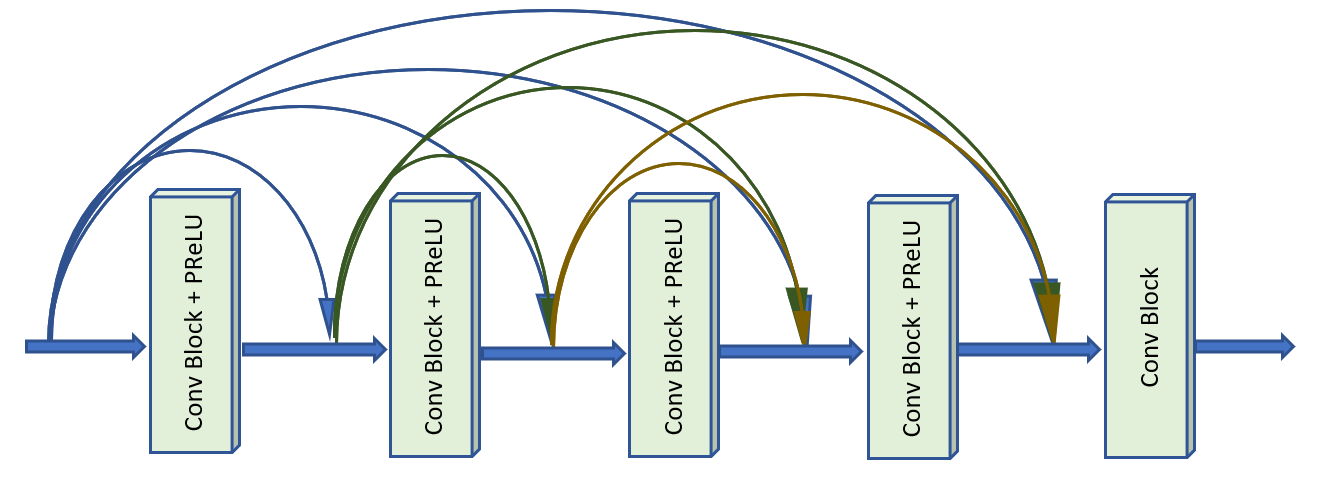}
\caption{Dense block from RRDB.}
\label{fig:denseBlock}
\vspace*{3mm}
\end{subfigure}
\caption{Internal diagram of RRDB (residual-in-residual dense blocks).}
\label{fig:RRDBDense}
\end{figure}

We use the generator architecture from ESRGAN \cite{ESRGAN}, where all batch normalization (BN) layers are removed, and RRDB is used. The overall architecture of generator $G$ is shown in figure \ref{fig:generator}, and the RRDB is depicted in figure \ref{fig:RRDBDense}. \par 
Inspired by the architecture of ESRGAN, we remove BN layers to increase the performance of the generator $G$ and to reduce the computational complexity. The authors of ESRGAN also state that the BN layers tend to introduce unpleasant artifacts and limit the generalization ability of the generator when the statistics of training and testing datasets differ significantly. \par 

We use RRDB as the basic blocks of the generator network $G$ that uses a multi-level residual network with dense connections. Those dense connections increase network capacity, and we also use residual scaling to prevent unstable conditions during the training phase \cite{ESRGAN}. We use the parametric rectified linear unit (PReLU) \cite{PRELU} for the dense blocks to learn the parameter with the other neural network parameters.
As discriminator ($D_{Ra}$), we employ a relativistic average discriminator similar to the work represented in \cite{ESRGAN}. \par

In equation \ref{eq:0} and \ref{eq:100}, the relativistic average discriminator is formulated for our architecture. Our generator $G$ depends on the discriminator $D_{Ra}$, and hence we briefly discuss the discriminator $D_{Ra}$ here and then, describe all details in section \ref{sec:discriminator}. The discriminator predicts the probability that a real image ($I_{HR}$) is relatively more realistic than a generated intermediate image ($I_{ISR}$). 

\begin{align}
    D_{Ra}(I_{HR}, I_{ISR}) &= \sigma(C(I_{HR}) - \E_{I_{ISR}}[C(I_{ISR})]) \xrightarrow{} 1 \hspace{3mm}\mbox{More Realistic than fake data?}
    \label{eq:0}\\
    D_{Ra}(I_{ISR},I_{HR}) &= \sigma(C(I_{ISR}) - \E_{I_{HR}}[C(I_{HR})]) \xrightarrow{} 0 \hspace{3mm}\mbox{Less realistic than real data?}
    \label{eq:100}
\end{align}

In equation \ref{eq:0} and \ref{eq:100}, $\sigma$, $C(\cdot)$ and $\E_{I_{ISR}}$ represents the sigmoid function, discriminator output and operation of calculating mean for all generated intermediate images in a mini-batch. The generated intermediate images are created by the generator where $I_{ISR} = G(I_{LR})$. It is evident from equation \ref{eq:1} that the adversarial loss of the generator contains both $I_{HR}$ and $I_{ISR}$ and hence, it benefits from the gradients of generated and ground truth images during the training process. The discriminator loss is depicted in equation \ref{eq:2}.
\begin{align}
\emph{L}_{G}^{Ra} &= -\E_{I_{HR}}[\log(1 - D_{Ra}(I_{HR}, I_{ISR}))] -\E_{I_{ISR}}[\log(D_{Ra}(I_{ISR}, I_{HR}))] \label{eq:1}\\
\emph{L}_{D}^{Ra} &= -\E_{I_{HR}}[\log(D_{Ra}(I_{HR}, I_{ISR}))] -\E_{I_{ISR}}[\log(1 - D_{Ra}(I_{ISR}, I_{HR}))]
\label{eq:2}
\end{align}
\par We use two more losses for generator G: one is perceptual loss ($\emph{L}_{percep}$), and another is content loss ($\emph{L}_{1}$) \cite{ESRGAN}. The perceptual loss is calculated using the feature map ($vgg_{fea}(\cdot)$) before the activation layers of a fine-tuned VGG19 \cite{vgg19} network, and the content loss calculates the 1-norm distance between $I_{ISR}$ and $I_{HR}$. Perceptual loss and content loss is shown in equation \ref{eq:3} and equation \ref{eq:4}.
\begin{align}
    \emph{L}_{percep} &= \E_{I_{LR}}||vgg_{fea}(G(I_{LR}) - vgg_{fea}(I_{HR})||_{1}
    \label{eq:3}\\
    \emph{L}_{1} &= \E_{I_{LR}}||G(I_{LR}) - I_{HR}||_{1} 
    \label{eq:4}
\end{align}
\subsubsection{Edge-Enhancement Network EEN}
\begin{figure}[H]
\centering
\includegraphics[width=15.6 cm, height=4.8cm]{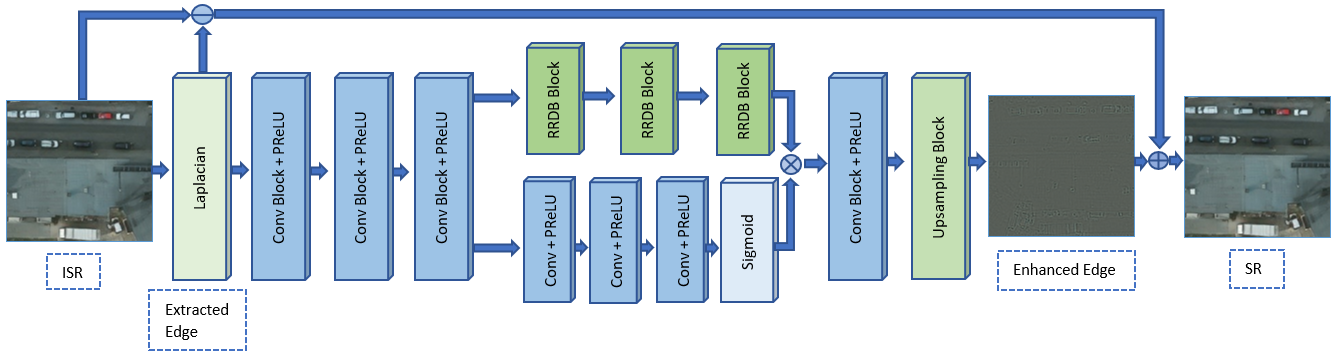}
\caption{Edge-enhancement network where input is an ISR (intermediate super-resolution) image and output is a SR (super-resolution) image.}
\label{fig:EEN}
\end{figure}

The EEN network removes noise and enhances the extracted edges from an image. An overview of the network is depicted in figure \ref{fig:EEN}. In the beginning, Laplacian operator \cite{laplacian_operator} is used to extract edges from the input image. After the edge information is extracted, it is passed through convolutional, RRDB, and upsampling blocks. There is a mask branch with sigmoid activation to remove edge noise as described in \cite{EEGAN}. Finally, the enhanced edges are added to the input images where the edges extracted by the Laplacian operator were subtracted. \par

The EEN network is similar to the edge-enhancement subnetwork proposed in \cite{EEGAN} with two improvements. First, we replace the dense blocks with RRDB. The RRDB shows improved performance according to ESRGAN \cite{ESRGAN}. Hence, we replace the dense block for improved performance of the EEN network. Secondly, we introduce a new loss term to improve the reconstruction of the edge information.

In \cite{EEGAN}, authors extracted the edge information from $I_{ISR}$ and enhanced the edges using an edge-enhancement subnetwork which is afterwards added to the edge-subtracted $I_{ISR}$. To train the network, \cite{EEGAN} proposed to use Charbonnier loss \cite{Charbonnier1994TwoDH} between the $I_{ISR}$ and $I_{HR}$. This function is called consistency loss for images ($\emph{L}_{img\_cst}$) and helps to get visually pleasant outputs with good edge information. However, sometimes the edges of some objects are distorted and produce some noises and consequently, do not give good edge information. Therefore, we introduce a consistency loss for the edges ($\emph{L}_{edge\_cst}$) as well. To compute $\emph{L}_{edge\_cst}$ we evaluate the Charbonnier loss between the extracted edges ($I_{edge\_SR}$) from ${I_{SR}}$ and the extracted edges ($I_{edge\_HR}$) from $I_{HR}$. The two consistency losses are depicted in equation \ref{eq:5} and equation \ref{eq:6} where $\rho(\cdot)$ is the Charbonnier penalty function \cite{charbonnier_penalty}. The total consistency loss is finally calculated for both images and edges by summing up the individual loss. The loss of our EEN is shown in equation \ref{eq:7}. 

\begin{align}
    \emph{L}_{img\_cst} &= \E_{I_{SR}} [\rho(I_{HR} - I_{SR})] \label{eq:5}\\
    \emph{L}_{edge\_cst} &= \E_{I_{edge\_SR}}[\rho(I_{edge\_HR} - I_{edge\_SR})] \label{eq:6}\\
    \emph{L}_{een} &= \emph{L}_{img\_cst} + \emph{L}_{edge\_cst} 
    \label{eq:7}
\end{align}
Finally, we get the overall loss for the generator module by adding the losses of the generator G and the EEN network. The overall loss for the generator module is shown in equation \ref{eq:8} where $\lambda_{1}$, $\lambda_{2}$, $\lambda_{3}$, and $\lambda_{4}$ are the weight parameters to balance different loss components. We empirically set the values as $\lambda_{1} = 1$, $\lambda_{2} = .001$, $\lambda_{3} = .01$, and $\lambda_{4} = 5$. 

\begin{equation}
\label{eq:8}
    \emph{L}_{G\_een} = \lambda_{1}\emph{L}_{percep} + \lambda_{2}\emph{L}_{G}^{Ra} + \lambda_{3}\emph{L}_{1} +  \lambda_{4}\emph{L}_{een}
\end{equation}

\subsection{Discriminator}
\label{sec:discriminator}
As described in the previous section, we use the relativistic discriminator $D_{Ra}$ for training the generator $G$. The architecture of the discriminator is taken from ESRGAN \cite{ESRGAN} which employs the VGG-19 \cite{vgg19} architecture. 
We use Faster R-CNN \cite{FRCNN} and SSD \cite{SSD} for our detector networks. The discriminator ($D_{Ra}$) and the detector network jointly act as discriminator for the generator module. We briefly describe these two detectors in the next two sections. \par
\subsubsection{Faster R-CNN}
The Faster R-CNN \cite{FRCNN} is a two-stage object detector and contains two networks: a region proposal network (RPN) to generate region proposals from an image and another network to detect objects from these proposals. In addition, the second network also tries to fit the bounding boxes around the detected objects.\par
The task of the RPN is to return image regions that have a high probability of containing an object. The RPN network uses a backbone network such as VGG \cite{vgg19}, ResNet, or ResNet with feature pyramid network \cite{FPNresnet50}. These networks are used as feature extractors, and different types of feature extractors can be chosen based on their performance on public datasets. We use ResNet-50-FPN \cite{FPNresnet50} as a backbone network for our faster R-CNN. We use this network because it displayed a higher precision than VGG-19 and ResNet-50 without FPN (especially for small object detection) \cite{FPNresnet50}. Even though the use of a larger network might lead to a further performance improvement, we chose ResNet-50-FPN due to its comparably moderate hardware requirements and more efficient convergence times. \par
After the RPN, there are two branches for detection: a classifier and a regressor. The classification branch is responsible for classifying a proposal to a specific object, and the regression branch finds the accurate bounding box of the object. In our case, both datasets contain objects with only one class, and therefore, our classifier infers only two classes: the background class and the object class. 
 
\subsubsection{SSD}
The SSD \cite{SSD} is a single-shot multibox detector that detects objects in a single stage. Here, single-stage means that classification and localization are done in a single forward pass through the network. Like Faster R-CNN, SSD also has a feature extractor network, and different types of networks can be used. To serve the primary purpose of SSD, which is speed, we use VGG-16 \cite{vgg19} as a feature extractor network. After this network, SSD has several convolutional feature layers of decreasing sizes. This representation can seem like a pyramid representation of images at different scales. Therefore, the detection of objects happens in every layer, and finally, we get the object detection output as class values and coordinates of bounding boxes.

\subsubsection{Loss of the discriminator}
The relativistic discriminator loss ($\emph{L}_{D}^{Ra}$) is already described in the previous section and depicted in equation \ref{eq:2}. This loss is added to the detector loss to get the final discriminator loss. 
\par Both Faster R-CNN and SSD have similar regression/localization losses  but different classification losses. 
For regression/localization, both use smooth $L_{1}$ \cite{FRCNN} loss between detected and ground truth bounding box coordinates ($t_{*}$). Classification ($\emph{L}_{cls\_frcnn}$) and regression loss ($\emph{L}_{reg\_frcnn}$) and overall loss ($\emph{L}_{det\_frcnn}$) of Faster R-CNN are given in the following:
\begin{align}
    \emph{L}_{cls\_frcnn} &= \E_{I_{LR}}[-\log(Det_{cls\_frcnn}(G_{G\_een}(I_{LR})))] \label{eq:9}\\
    \emph{L}_{reg\_frcnn} &= \E_{I_{LR}}[smooth_{L1}(Det_{reg\_frcnn}(G_{G\_een}(I_{LR})), t_{*})] \label{eq:10}\\
    \emph{L}_{det\_frcnn} &= \emph{L}_{cls\_frcnn} + \lambda\emph{L}_{reg\_frcnn} \label{eq:11}
\end{align}

Here, $\lambda$ is used to balance the losses, and it is set to 1 empirically. $Det_{cls\_frcnn}$ and $Det_{reg\_frcnn}$ are the classifier and regressor for the Faster R-CNN. Classification ($\emph{L}_{cls\_ssd}$), regression loss ($\emph{L}_{reg\_ssd}$) and overall loss ($\emph{L}_{det\_ssd}$) of SSD are as following: 
\begin{align}
    \emph{L}_{cls\_ssd} &= \E_{I_{LR}}[-\log(softmax(Det_{cls\_ssd}(G_{G\_een}(I_{LR}))))] \label{eq:12}\\
    \emph{L}_{reg\_ssd} &= \E_{I_{LR}}[smooth_{L1}(Det_{reg\_ssd}(G_{G\_een}(I_{LR})), t_{*})] \label{eq:13}\\
    \emph{L}_{det\_ssd} &= \emph{L}_{cls\_ssd} + \alpha\emph{L}_{reg\_ssd} \label{eq:14}
\end{align}
Here, $\alpha$ is used to balance the losses, and it is set to 1 empirically. $Det_{cls\_ssd}$ and $Det_{reg\_ssd}$ are the classifier and regressor for the SSD. 

\subsection{Training}
Our architecture can be trained in separate steps or jointly in an end-to-end way. We discuss the details of these two types of training in the next two sections.
\subsubsection{Separate Training}
In separate training, we train the SR network (generator module and discriminator $D_{Ra}$) and the detector separately. Detector loss is not backpropagated to the generator module. Therefore, the generator is not aware of the detector and thus, it only gets feedback from the discriminator $D_{Ra}$. For example, in equation \ref{eq:9}, no error is backpropagated to the $G_{G\_een}$ network (the network is detached during the calculation of the detector loss) while calculating the loss $\emph{L}_{cls\_frcnn}$. 
\subsubsection{End-to-End Training}
In end-to-end training, we train the whole architecture end-to-end that means the detector loss is backpropagated to the generator module. Therefore, the generator module revceives gradients from both detector and discriminator $D_{Ra}$. We get the final discriminator loss ($\emph{L}_{D_\_det}$) as following:
\begin{equation}
    \emph{L}_{D_\_det} = \emph{L}_{D}^{Ra} + \eta\emph{L}_{det}
    \label{eq:15}
\end{equation}
Here, $\eta$ is the parameter to balance the contribution of the detector loss and we empirically set it to 1. Finally, we get an overall loss ($\emph{L}_{overall}$) for our architecture as follows.
\begin{equation}
    \label{eq:16}
    \emph{L}_{overall} = \emph{L}_{G\_een} + \emph{L}_{D_\_det}
\end{equation}
\par

\section{Experiments}
As mentioned above, we trained our architecture separately and in an end-to-end manner. For separate training, we first trained the SR network until convergence and then trained the detector networks based on the SR images. For end-to-end training, we also employed separate training as pre-training step for weight initialization. Afterwards SR and object detection networks were jointly trained, i.e., the gradients from the the object detector were propagated into the generator network.
\par
In the training process, the learning rate was set to 0.0001 and halved after every 50$k$ iterations. The batch size was set to 5. 
We used Adam \cite{adam} as optimizer with $\beta_{1} = 0.9$, $\beta_{2} = 0.999$ and updated the whole architecture weights until convergence. We used 23 RRDB blocks for the generator $G$ and 5 RRDB blocks for the EEN network. We implemented our architecture with the PyTorch framework \cite{pytorch} and trained/tested using two NVIDIA Titan X GPUs. The end-to-end training with COWC took 96 hours for 200 epochs. The average inference speed using faster R-CNN was approximately 4 images/second and 7 images/second for SSD. Our implementation can be found in GitHub \cite{Filter_Enhance_Detect}. 
\label{sec:results}
\subsection{Datasets}
\subsubsection{Cars Overhead with Context Dataset}

Cars overhead with context (COWC) dataset \cite{cowc} contains 15 cm (one pixel cover 15 cm distance at ground level) satellite images from six different regions. The dataset contains a large number of unique cars and covers regions from Toronto in Canada, Selwyn in New Zealand, Potsdam and Vaihingen in Germany, Columbus and Utah in the United States. Out of these six regions, we used the dataset from Toronto and Potsdam. Therefore, when we refer to the COWC dataset, we refer to the dataset from these two regions. There are 12651 cars in our selected dataset. The dataset contains only RGB images, and we used these images for training and testing. \par

We used 256-by-256 image tiles, and every image tile contains at least one car. The average length of a car was between 24 to 48 pixels, and the width was between 10 to 20 pixels. Therefore, the area of a car was between 240 to 960 pixels, which can be considered as a small object relative to the other large satellite objects. We used bi-cubic downsampling to generate LR images from the COWC dataset. The downscale factor was 4x, and therefore, we had 64 pixels to 64 pixels size for LR images. We had a text file associated with each image tile containing the coordinates of the bounding box for each car. \par

\begin{figure}[H]
\centering
\renewcommand{\arraystretch}{0.01}
\begin{tabular}{| @{} c @{} | @{} c @{} | @{} c @{}|}
\hline
\includegraphics[width = 1.5in]{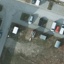} &
\includegraphics[width = 1.5 in]{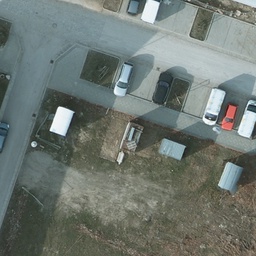} &
\includegraphics[width = 1.5in]{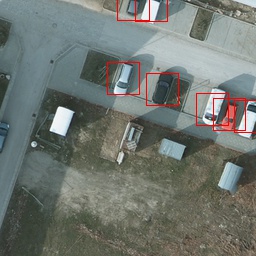} \\
\hline
\includegraphics[width = 1.5in]{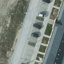} &
\includegraphics[width = 1.5 in]{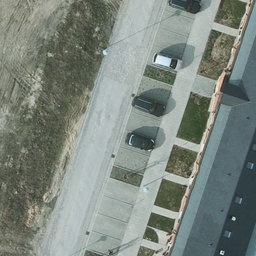} &
\includegraphics[width = 1.5in]{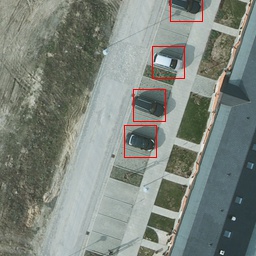} \\
\hline
\includegraphics[width = 1.5in]{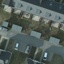} &
\includegraphics[width = 1.5 in]{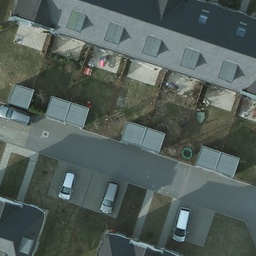} &
\includegraphics[width = 1.5in]{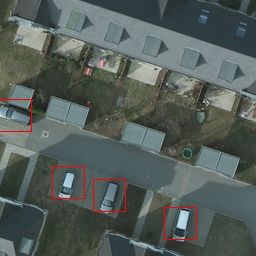} \\
\hline
\arrayrulecolor{white}
\hline
\hline
\multicolumn{1}{c}{(a) LR image} & \multicolumn{1}{c}{(b) HR image} & \multicolumn{1}{c}{(c) GT image}
\end{tabular}
\caption{COWC (car overhead with context) dataset: LR-HR (low-resolution and high-resolution) image pairs are shown in (a) and (b) and GT (ground truth) images with bounding boxes for cars are in (c).}
\label{fig:cowc}
\end{figure}

Our experiments considered the dataset having only one class, car, and did not consider any other type of object. Figure \ref{fig:cowc} shows examples from the COWC dataset. We experimented with a total of 3340 tiles for training and testing. Our train/test split was 80\%/20\%, and the training set was further divided into a training and a validation set by an 80\% to 20\% ratio. We trained our end-to-end architecture with an augmented training dataset with random horizontal flips and ninety-degree rotations.

\subsubsection{Oil and Gas Storage Tank Dataset}
The oil and gas storage tank (OGST) dataset has been complied in Alberta Geological Survey (AGS) \cite{AGS}, a branch of the Alberta Energy Regulatory (AER) \cite{AER}. AGS provides geoscience information and support to AER’s regulatory functions on energy developments to be carried out in a manner to ensure public and environmental safety. To assist AER with sustainable land management and compliance assurance \cite{ags_subir}, AGS is utilizing remote sensing imagery for identifying the number of oil and gas storage tanks inside well pad footprints in Alberta. \par

While the SPOT-6 satellite imagery at 1.5 m pixel resolution provided by the AGS has sufficient quality and details for many regulatory functions, it is difficult to detect small objects within well pads, e.g., oil and gas storage tanks with ordinary object detection methods. The diameter of a typical storage tank is about 3 m and their placements are usually vertical and side-by-side with less than 2 m. To train our architecture for this use-case, we needed a dataset for providing pairs of low and high-resolution images. Therefore, we have created the OGST dataset using free imagery from the Bing map \cite{bing}.

\par
The OGST dataset contains 30 cm resolution remote sensing images (RGB) from the Cold Lake Oil Sands region of Alberta, Canada where there is a high level of oil and gas activities and concentration of well pad footprints. The dataset contains 1671 oil and gas storage tanks from this area.
\par
\begin{figure}[H]
\centering
\renewcommand{\arraystretch}{0.01}
\begin{tabular}{| @{} c @{} | @{} c @{} | @{} c @{}|}
\hline
\includegraphics[width = 1.5in]{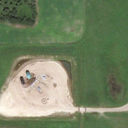} &
\includegraphics[width = 1.5 in]{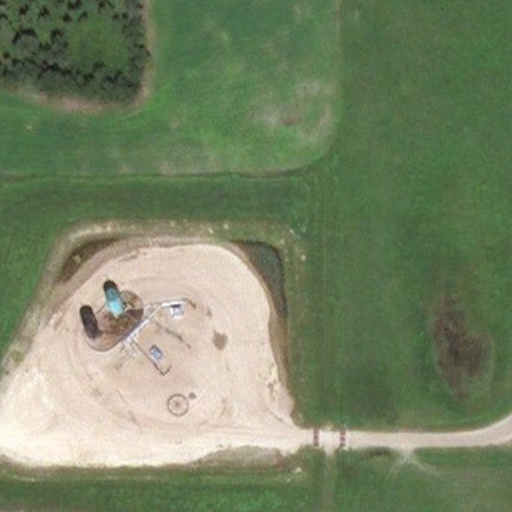} &
\includegraphics[width = 1.5in]{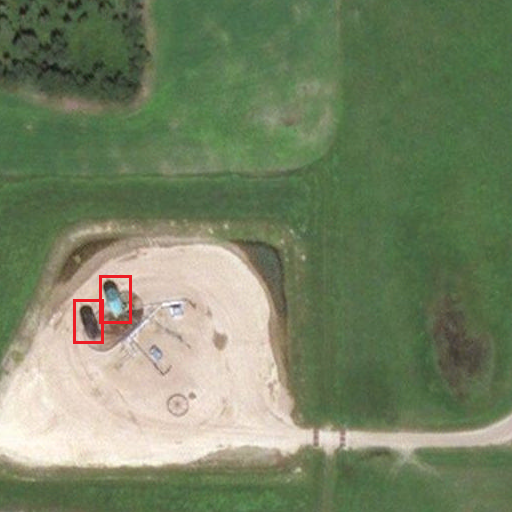} \\
\hline
\includegraphics[width = 1.5in]{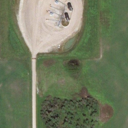} &
\includegraphics[width = 1.5 in]{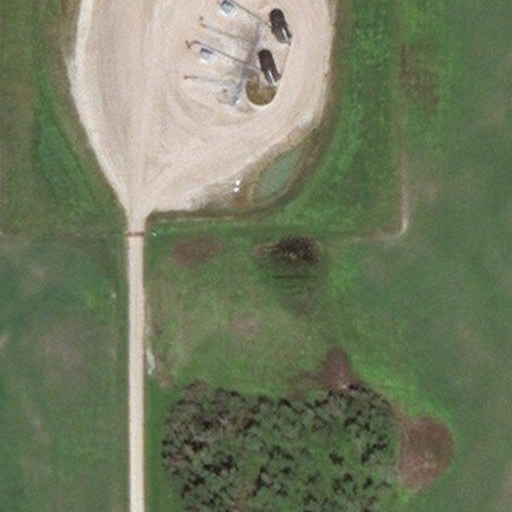} &
\includegraphics[width = 1.5in]{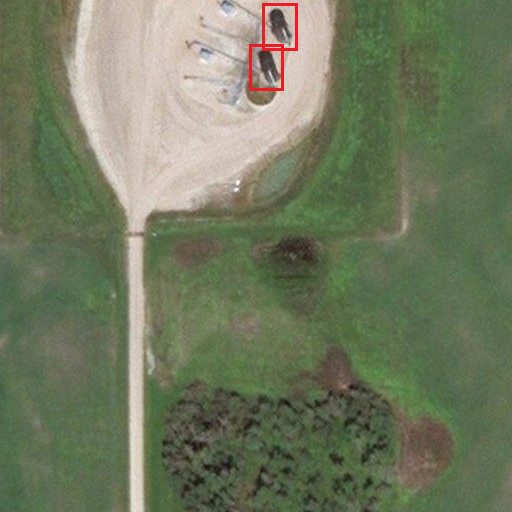} \\
\hline
\includegraphics[width = 1.5in]{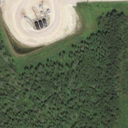} &
\includegraphics[width = 1.5 in]{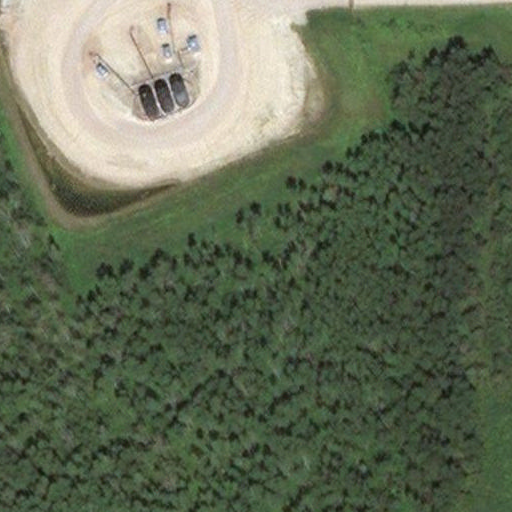} &
\includegraphics[width = 1.5in]{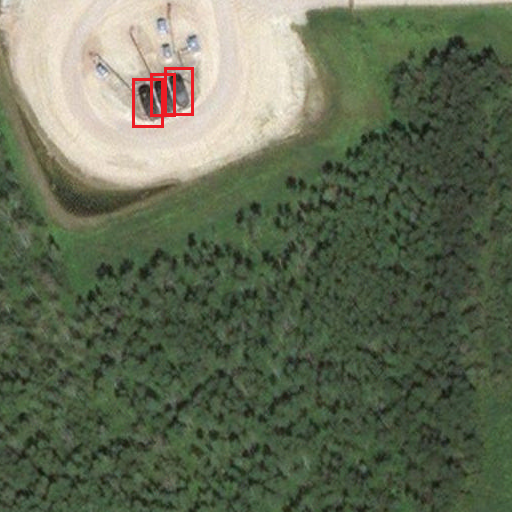} \\
\hline
\arrayrulecolor{white}
\hline
\hline
\multicolumn{1}{c}{(a) LR image} & \multicolumn{1}{c}{(b) HR image} & \multicolumn{1}{c}{(c) GT image}
\end{tabular}
\caption{OGST (oil and gas storage tank) dataset: LR-HR (low-resolution and high-resolution) image pairs are shown in (a) and (b) and GT (ground truth) images with bounding boxes for oil and gas storage tanks are in (c).}
\label{fig:TankDataset}
\end{figure}

We used 512-by-512 image tiles, and there was no image without any oil and gas storage tank in our experiment. The average area covered by an individual tank was between 800 to 1600 pixels. Some industrial tanks were large, but most of the tanks covered small regions on the imagery. We downscaled the HR images using bi-cubic downsampling with the factor of 4x, and therefore, we got a LR tile of size 128-by-128 pixels. Every image tile was associated with a text file containing the coordinates of the bounding boxes for the tanks on a tile. We have showed examples from the OGST dataset in figure \ref{fig:TankDataset}. \par

\par

As with the COWC dataset, our experiments considered one unique class here, tank, and we had a total of 760 tiles for training and testing. We used a 90\%/10\% split for our train/test data. The training data was further divided by 90\%/10\% for the train/validation split. The percentage of training data was higher here compared to the previous dataset to increase the training data because of the smaller size of the dataset. The dataset is available at \cite{Filter_Enhance_Detect}.

\subsection{Evaluation Metrics for Detection}
\label{ssec:Eval_mat}
We obtained our detection output as bounding boxes with associated classes. To evaluate our results, we used average precision (AP), and calculated intersection over union (IoU), precision,
and recall for obtaining AP.
\par
We denote the set of correctly detected objects as true positives (TP) and the set of falsely detected objects of false positives (FP). The precision is now the ratio between the number of TPs relative to all predicted objects:

\begin{equation}
precision=\frac{|TP|}{|TP|+|FP|}
\label{eq:17}
\end{equation}

We denote the set of objects which are not detected by the detector as false negatives (FN). Then, the recall is defined as the ratio of detected objects (TP) relative to the number of all objects in the data set:

\begin{equation}
Recall = \frac{|TP|}{|TP|+|FN|}
\label{eq:18}
\end{equation}

\par
To measure the localization error of predicted bounding boxes, IoU computes the overlap between two bounding boxes: the detected and the ground truth box. If we take all the boxes that have an IoU $\geq \tau$ as TP and consider all other detections as FP, then we get the precision at $\tau$ IoU. 
If we now vary $\tau$ from 0.5 to 0.95 IoU with a step size of 0.05, we receive ten different precision values which can be combined into the average precision (AP) at IoU=0.5:0.95 \cite{FRCNN}.
Let us note that in the case of multi-class classification, we would need to compute the AP for object each class separately. To receive a single performance measure for object detection, the mean AP (mAP) is computed which is the most common performance measure for object detection quality. \par
In this paper, both of our datasets only contain single class, and hence, we used AP as our evaluation metric. We mainly showed the results of AP at IoU=0.5:0.95 as our method performed increasingly better compared to other models when we increased the IoU values for AP calculation. We show this trend in section \ref{sssec:APIoU}.

\subsection{Results}

\subsubsection{Detection without Super-Resolution}
We ran the two detectors to document the object detection performance on both LR and HR images. We used SSD with vgg16 \cite{vgg19} network and Faster R-CNN (FRCNN) with ResNet-50-FPN \cite{FPNresnet50} detector. We trained the two models with both HR and 4x-downscaled LR images. Testing was also done with both HR and LR images.
\par

{\renewcommand{\arraystretch}{1.5}
\begin{table}[H]
\renewcommand\thetable{1}
\centering
\caption{Detection on LR (low-resolution) and HR (high-resolution) images without using super-resolution. Detectors are trained with both LR and HR images and AP (average precision) values are calculated using 10 different IoUs (intersection over union).}

\begin{tabular}{ | >{\centering\arraybackslash}m{3cm}| >{\centering\arraybackslash}m{3cm}| >{\centering\arraybackslash}m{3cm}|
>{\centering\arraybackslash}m{3cm}|} 

\hline
\rowcolor{gray!10}
\textbf{Model} & \textbf{Training image resolution - Test image  resolution} & \textbf{COWC dataset \hspace{3mm}(Test Results) \hspace{6mm} (AP at IoU=0.5:0.95) (single class - 15 cm)} & \textbf{OGST dataset \hspace{15mm}(Test Results)\hspace{6mm} (AP at IoU=0.5:0.95) (single class - 30 cm)}\\
\hline
\multirow{2}{*}{SSD} & LR - LR & 61.9\% & 76.5\%\\ 
\cline{2-4}
& HR - LR & 58\% & 75.3\%\\ 
\hline
\multirow{2}{*}{FRCNN} & LR - LR & 64\% & 77.3\%\\ 
\cline{2-4} 
& HR - LR & 59.7\% & 75\% \\ 
\hline
SSD-RFB & LR - LR & 63.1\% & 76.7\% \\
\hline
\hline
\hline
SSD & HR - HR & 94.1\% & 82.5\% \\ \hline
FRCNN & HR - HR & 98\% & 84.9\%  \\ 
\hline
\end{tabular}
\label{table:1}
\end{table}
}

In table \ref{table:1}, we show the results of the detection performance of the detectors with different train/test combinations. When we only used LR images for both training and testing, we observed 64\% AP for Faster R-CNN. When training on HR images and testing with LR images, the accuracy dropped for both detectors. We also added detection results (using LR images for training/testing) for both the datasets using SSD with RFB modules (SSD-RFB) \cite{RFBNet}, where accuracy slightly increased from the base SSD. \par
The last two rows in table \ref{table:1} depict the accuracy of both detectors when training and testing on HR images. We have achieved up to 98\% AP with the Faster R-CNN detector. This, shows the large impact of the resolution to the object detection quality and sets a natural upper bound on how close a SR-based method can get when working on LR images. In the next sections, we demonstrate that our approaches considerably improve the detection rate on LR imagery and get astonishingly close to the performance of directly working on HR imagery.

\subsubsection{Separate Training with Super-Resolution}

{\renewcommand{\arraystretch}{1.5}
\begin{table}[H]
\renewcommand\thetable{2}
\centering
\caption{Detection on SR (super-resolution) images with separately trained SR network. Detectors are trained with both SR and HR (high-resolution) images and AP (average precision) values are calculated using 10 different IoUs (intersection over union).}
\begin{tabular}{ | >{\centering\arraybackslash}m{3cm}| >{\centering\arraybackslash}m{3cm}| >{\centering\arraybackslash}m{3cm}|
>{\centering\arraybackslash}m{3cm}|} 

\hline
\rowcolor{gray!10}
\textbf{Model} & \textbf{Training image resolution - Test image  resolution} & \textbf{COWC Dataset \hspace{3mm}(Test Results) \hspace{6mm} (AP at IoU=0.5:0.95) (single class - 15 cm)} & \textbf{OGST Dataset \hspace{15mm}(Test Results) \hspace{6mm}  (AP at IoU=0.5:0.95)  (single class - 30 cm)} \\
\hline
\multirow{2}{*}{Bicubic + SSD} & SR - SR & 72.1\% & 77.6\% \\ 
\cline{2-4}
& HR - SR & 58.3\% & 76\%\\ 
\hline
\multirow{2}{*}{Bicubic + FRCNN} & SR - SR & 76.8\% & 78.5\% \\ 
\cline{2-4} 
& HR - SR & 61.5\%  & 77.1\% \\ 
\hline
\multirow{2}{*}{EESRGAN + SSD} & SR - SR & 86\% & 80.2\% \\ 
\cline{2-4}
& HR - SR & 83.1\% & 79.4\%\\ 
\hline
\multirow{2}{*}{EESRGAN + FRCNN} & SR - SR & \textbf{93.6\%} & \textbf{81.4\%}\\ 
\cline{2-4} 
& HR - SR & 92.9\%  & 80.6\%\\ 
\hline
\multirow{2}{*}{ESRGAN + SSD} & SR - SR & 85.8\% & 80.2\%\\ 
\cline{2-4}
& HR - SR & 82.5\%  & 78.9\% \\ 
\hline
\multirow{2}{*}{ESRGAN + FRCNN} & SR - SR & 92.5\% & 81.1\%\\ 
\cline{2-4}
& HR - SR & 91.8\%  & 79.3\% \\ 
\hline
\multirow{2}{*}{EEGAN + SSD} & SR - SR & 86.1\% &  79.1\%\\ 
\cline{2-4}
& HR - SR & 83.3\% & 77.5\% \\ 
\hline
\multirow{2}{*}{EEGAN + FRCNN} & SR - SR &  92\% & 79.9\% \\ 
\cline{2-4} 
& HR - SR & 91.1\%  & 77.9\% \\ 
\hline
\end{tabular}
\label{table:2}
\end{table}
}

In this experiment, we created 4x upsampled images from the LR input images using bicubic upsampling and different SR methods. Let us note that no training was needed for applying bicubic upsampling since it is a parameter free function. We used the SR images as test data for two types of detectors. We compared three GAN architectures for generating SR images, our new EESRGAN architecture, ESRGAN \cite{ESRGAN} and EEGAN \cite{EEGAN}. Each network was trained separately on the training set before the object detector was trained. For the evaluation, we again compared detectors being trained on the SR images from the particular architecture and detectors being directly trained on the HR images.
\par

In table \ref{table:2}, the detection output of the different combinations of SR methods and detectors is shown with the different combinations of train/test pairs. As can be seen, our new EESRGAN architecture displayed the best results already getting close to the detection rates which could be observed when working with HR images only. However, after training EESRGAN can be directly applied to LR imagery where no HR data is available and still achieved very good results.
Furthermore, we could observe that other SR methods EEGAN and ESRGAN have already improved the AP considerably when used for preprocessing of LR images. However, for both data sets, EESRGAN have outperformed the other two methods. 

\subsubsection{End-to-End Training with Super-Resolution}
We trained our EESRGAN network and detectors end-to-end for this experiment. The discriminator ($D_{Ra}$), and the detectors jointly acted as a discriminator for the entire architecture. Detector loss was backpropagated to the SR network, and therefore, the loss contributed to the enhancement of LR images. At training time, LR-HR image pairs were used to train the EEGAN part, and then the generated SR images were sent to the detector for training. At test time, only the LR images were fed to the network. Our architecture first generated a SR image of the LR input before object detection was performed. 
{\renewcommand{\arraystretch}{1.5}
\begin{table}[H]
\renewcommand\thetable{3}
\centering
\caption{Detection with end-to-end SR (super-resolution) network. Detectors are trained with SR images and AP (average precision) values are calculated using 10 different IoUs (intersection over union).}
\begin{tabular}{ | >{\centering\arraybackslash}m{3cm}| >{\centering\arraybackslash}m{3cm}| >{\centering\arraybackslash}m{3cm}|
>{\centering\arraybackslash}m{3cm}|} 

\hline
\rowcolor{gray!10}
\textbf{Model} & \textbf{Training image resolution - Test image  resolution} & \textbf{COWC Dataset \hspace{3mm}(Test Results) \hspace{6mm} (AP at IoU=0.5:0.95) (single class - 15 cm)} & \textbf{OGST Dataset \hspace{15mm}(Test Results) \hspace{6mm} (AP at IoU=0.5:0.95) (single class - 30 cm)} \\
\hline
EESRGAN + SSD & SR - SR & 89.3\% & 81.8\% \\ 
\hline
EESRGAN + FRCNN & SR - SR & \textbf{95.5\%} & \textbf{83.2\%} \\ 
\hline
ESRGAN + SSD & SR - SR & 88.5\% & 81.1\% \\ 
\hline
ESRGAN + FRCNN & SR - SR & 93.6\% & 82\% \\ 
\hline
EEGAN + SSD & SR - SR & 88.1\% & 80.8\% \\ 
\hline
EEGAN + FRCNN & SR - SR & 93.1\% & 81.3\% \\ 
\hline
\end{tabular}
\label{table:3}
\end{table}
}

We also compared our results with different architectures. We used ESRGAN \cite{ESRGAN} and EEGAN \cite{EEGAN} with the detectors for comparison. Table \ref{table:3} clearly shows that our method delivers superior results compared to others.

\subsubsection{AP versus IoU curve}
\label{sssec:APIoU}
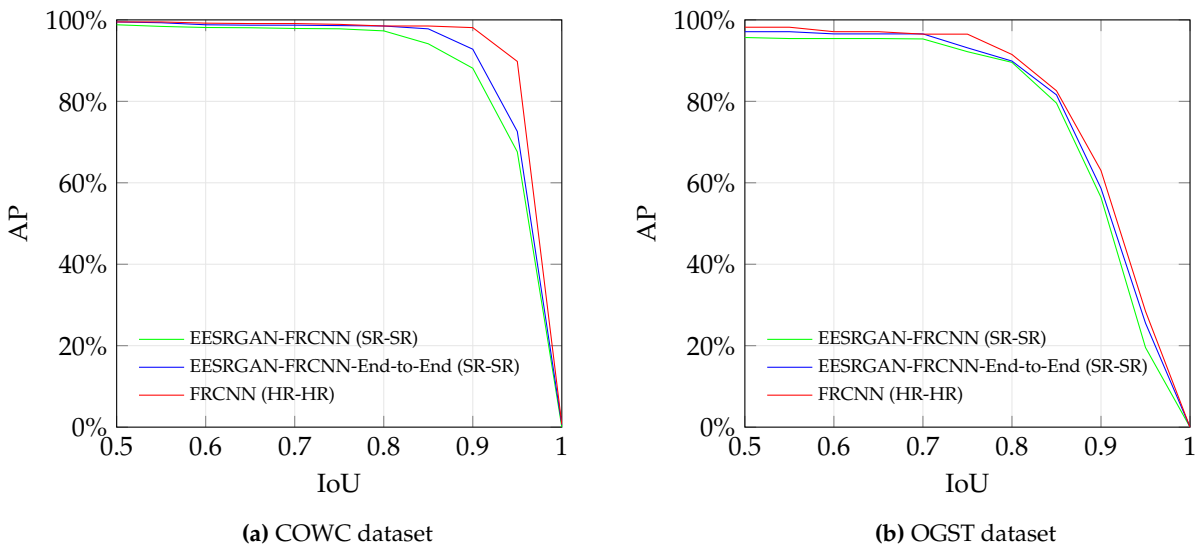
\begin{figure}[H]
\captionsetup[subfigure]{oneside,margin={2.5cm,0cm}}
\begin{adjustbox}{max width=\textwidth}
\centering
\begin{subfigure}[b]{.4\textwidth}
\centering
\begin{tikzpicture}
\begin{axis}[
    width=7.5cm,
    height=7cm,
    xlabel={IoU},
    ylabel={AP},
    xmin=0.5, xmax=1,
    ymin=0, ymax=100,
    xtick={0.50,0.60,0.70,0.80,0.90, 1.0},
    yticklabel={$\pgfmathprintnumber{\tick}\%$},
    legend pos=north west,
    legend cell align=left,
    grid=both,
    grid style={line width=.1pt, draw=gray!20},
    legend style={legend pos=south west, draw=none, fill=none, font=\fontsize{7}{5}\selectfont}
]
    
\addplot[
    color=green
    ]
    coordinates {
    (0.50,98.8)(.55,98.4)(0.60,98.13)(0.65,98.07)(0.70,97.89)(0.75,97.81)(0.80,97.3)(0.85,94.13)(0.90,88.13)(0.95,67.6)(1.0,0)
    };
    \addlegendentry{EESRGAN-FRCNN (SR-SR)}

\addplot[
    color=blue
    ]
    coordinates {
    (0.50,99.5)(.55,99.3)(0.60,98.8)(0.65,98.7)(0.70,98.7)(0.75,98.6)(0.80,98.5)(0.85,97.8)(0.90,92.8)(0.95,72.6)(1.0,0.59)
    };
    \addlegendentry{EESRGAN-FRCNN-End-to-End (SR-SR)}
    
    \addplot[
    color=red
    ]
    coordinates {
    (0.50,99.5)(.55,99.5)(0.60,99.2)(0.65,99.1)(0.70,99.1)(0.75,98.9)(0.80,98.5)(0.85,98.5)(0.90,98.1)(0.95,89.8)(1.0,0.67)
    };
    \addlegendentry{FRCNN (HR-HR)}
    
\end{axis}
\end{tikzpicture}
\caption{COWC dataset}
\end{subfigure}
\hfill
\hspace{17mm}
\begin{subfigure}[b]{.4\textwidth}
\centering
\begin{tikzpicture}
\begin{axis}[
    width=7.5cm,
    height=7cm,
    xlabel={IoU},
    ylabel={AP},
    xmin=0.5, xmax=1.0,
    ymin=0, ymax=100,
    xtick={0.50,0.60,0.70,0.80,0.90, 1.0},
    yticklabel={$\pgfmathprintnumber{\tick}\%$},
    legend pos=north west,
    legend cell align=left,
    grid=both,
    grid style={line width=.1pt, draw=gray!20},
    legend style={legend pos=south west, draw=none, fill=none, font=\fontsize{7}{5}\selectfont}
]
    
\addplot[
    color=green
    ]
    coordinates {
    (0.50,95.67)(.55,95.42)(0.60,95.42)(0.65,95.42)(0.70,95.32)(0.75,92.19)(0.80,89.57)(0.85,79.57)(0.90,56.41)(0.95,19.54)(1.0,0)
    };
    \addlegendentry{EESRGAN-FRCNN (SR-SR)}

\addplot[
    color=blue
    ]
    coordinates {
    (0.50,97.1)(.55,97.1)(0.60,96.56)(0.65,96.56)(0.70,96.56)(0.75,93.14)(0.80,89.89)(0.85,81.59)(0.90,58.64)(0.95,25.54)(1.0,0)
    };
    \addlegendentry{EESRGAN-FRCNN-End-to-End (SR-SR)}
    
\addplot[
    color=red
    ]
    coordinates {
    (0.50,98.2)(.55,98.2)(0.60,97.1)(0.65,97.1)(0.70,96.5)(0.75,96.5)(0.80,91.5)(0.85,82.6)(0.90,63)(0.95,28.45)(1.0,0)
    };
    \addlegendentry{FRCNN (HR-HR)}
    
\end{axis}
\end{tikzpicture}
\caption{OGST dataset}
\end{subfigure}
\end{adjustbox}
\vspace{0.1cm}
\caption{AP-IoU (average precision-intersection over union) curves for the datasets. Plotted results show the detection performance of standalone faster R-CNN on HR (high-resolution) images and our proposed method (with and without end-to-end training) on SR (super-resolution) images.}
\label{fig:AP_IoU}
\end{figure}

We have calculated the AP values on different IoUs. In figure \ref{fig:AP_IoU}, we plot the AP versus IoU curves for our datasets. The performance of EESRGAN-FRCNN, end-to-end EESRGAN-FRCNN, and FRCNN is shown in the figure. The end-to-end EESRGAN-FRCNN network has performed better than the separately trained network. The difference is most evident for the higher IoUs on the COWC dataset. \par

Our results indicate excellent performance compared to the highest possible AP values obtained from standalone FRCNN (trained and tested on HR images) \par
The OGST dataset has displayed less performance variation compared to the COWC dataset. The object size of the OGST dataset is larger than that of the COWC dataset. Therefore, the performance difference was not similar to the COWC dataset when we compared between standalone FRCNN and our method on the OGST dataset. To conclude, training our new architecture in an end-to-end manner has displayed an improvement for both the datasets. 

\captionsetup[figure]{width=.9\linewidth}

\noindent

\subsubsection{Precision versus Recall}
\label{sssec:PreRec}

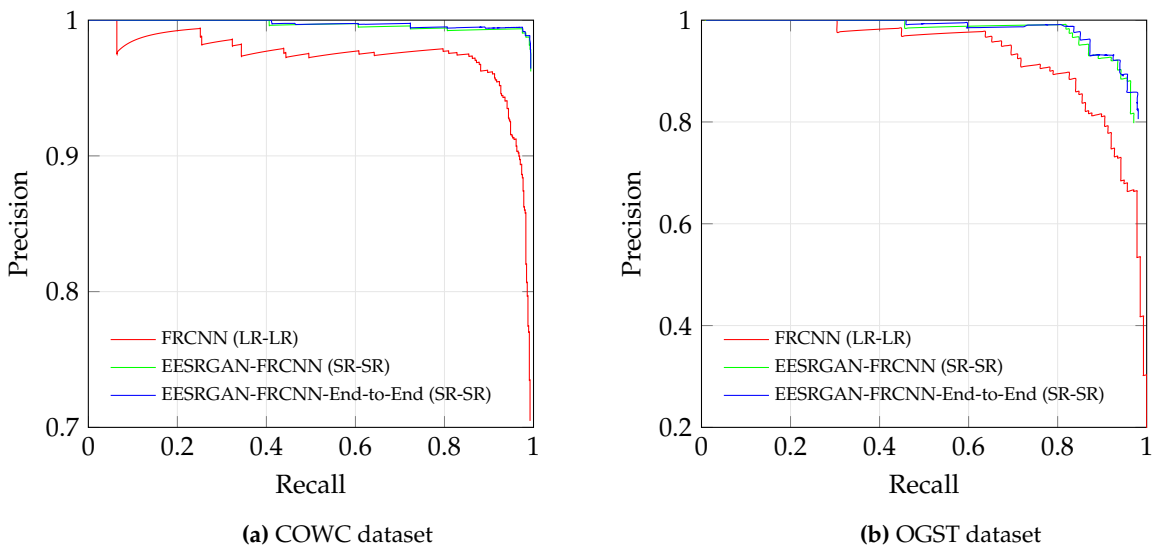
\begin{figure}[H]
\captionsetup[subfigure]{oneside,margin={2.5cm,0cm}}
\begin{adjustbox}{max width=\textwidth}
\centering
\begin{subfigure}[b]{.4\textwidth}
\centering
\begin{tikzpicture}
\begin{axis}[
    width=7.5cm,
    height=7cm,
    xlabel={Recall},
    ylabel={Precision},
    xmin=0, xmax=1,
    ymin=0.7, ymax=1,
    xtick={0,0.2,0.4,0.6,0.8,1.0},
    ytick={0.1,0.2,0.3,0.4,0.5,0.6,0.7,0.8,0.9,1.0},
    legend pos=north west,
    legend cell align=left,
    grid=both,
    grid style={line width=.1pt, draw=gray!20},
    legend style={legend pos=south west, draw=none, fill=none, font=\fontsize{7}{5}\selectfont}
]

\addplot[smooth,red] table[
    x=recall, 
    y=precision, 
    col sep=comma
    ]
    {Definitions/data/precision_recall_LR_LR_cowc.csv};
    \addlegendentry{FRCNN (LR-LR)}
    
\addplot[smooth,green] table[
    x=recall, 
    y=precision, 
    col sep=comma
    ]
    {Definitions/data/precision_recall_SR_SR_cowc.csv};
    \addlegendentry{EESRGAN-FRCNN (SR-SR)}
    
\addplot[smooth,blue] table[
    x=recall, 
    y=precision, 
    col sep=comma
    ]
    {Definitions/data/precision_recall_SR_SR_cars_end_end.csv};
    \addlegendentry{EESRGAN-FRCNN-End-to-End (SR-SR)}
    
\end{axis}
\end{tikzpicture}
\caption{COWC dataset}
\label{fig:precision-recall-a}
\end{subfigure}
\hfill
\hspace{15mm}
\begin{subfigure}[b]{.4\textwidth}
\centering
\begin{tikzpicture}
\begin{axis}[
    width=7.5cm,
    height=7cm,
    xlabel={Recall},
    ylabel={Precision},
    xmin=0, xmax=1,
    ymin=0.2, ymax=1,
    xtick={0,0.2,0.4,0.6,0.8,1.0},
    ytick={0.2,0.4,0.6,0.8,1.0},
    legend pos=north west,
    legend cell align=left,
    grid=both,
    grid style={line width=.1pt, draw=gray!20},
    legend style={legend pos=south west, draw=none, fill=none, font=\fontsize{7}{5}\selectfont}
]

\addplot[smooth,red] table[
    x=recall, 
    y=precision, 
    col sep=comma
    ]
    {Definitions/data/precision_recall_LR_LR_tank.csv};
    \addlegendentry{FRCNN (LR-LR)}
    
\addplot[smooth,green] table[
    x=recall, 
    y=precision, 
    col sep=comma
    ]
    {Definitions/data/precision_recall_SR_SR_tank.csv};
    \addlegendentry{EESRGAN-FRCNN (SR-SR)}
    
\addplot[smooth,blue] table[
    x=recall, 
    y=precision, 
    col sep=comma
    ]
    {Definitions/data/precision_recall_SR_SR_tank_end_end.csv};
    \addlegendentry{EESRGAN-FRCNN-End-to-End (SR-SR)}
    
\end{axis}
\end{tikzpicture}
\caption{OGST dataset}
\label{fig:precision-recall-b}
\end{subfigure}
\end{adjustbox}
\vspace{0.1cm}
\caption{Precision-recall curve for the datasets. Plotted results show the detection performance of standalone faster R-CNN on LR (low-resolution) images and our proposed method (with and without end-to-end training) on SR (super-resolution) images.}
\label{fig:precision-recall}
\end{figure}

In figure \ref{fig:precision-recall}, precision-recall curves are shown for both of our datasets. The precision-recall curve for COWC dataset is depicted in  \ref{fig:precision-recall-a} and \ref{fig:precision-recall-b} represents the curve for OGST dataset. For each dataset, we plot the curves for standalone faster R-CNN with LR training/testing images, and our method with/without end-to-end training. We used IoU=0.5 to calculate precision and recall.
\par

\captionsetup[figure]{width=.9\linewidth}

The precision-recall curves for both datasets show that our method has higher precision values in higher recall values compared to the standalone faster R-CNN models. Our models with end-to-end training performed better than our models without the end-to-end training. In particular, the end-to-end models have detected more than 99\% of the cars with 96\% AP in the COWC dataset. For the OGST dataset, our end-to-end models have detected more than 81\% of the cars with 97\% AP.

\subsubsection{Effects of Dataset Size}
\begin{figure}[H]
\captionsetup[subfigure]{oneside,margin={2.5cm,0cm}}
\begin{adjustbox}{max width=\textwidth}
\centering
\begin{subfigure}[b]{.4\textwidth}
\centering
\centering
\begin{tikzpicture}
\begin{axis}[
    width=7.5cm,
    height=7cm,
    xlabel={Dataset Size ($\cdot10^4$)},
    ylabel={AP},
    xmin=0, xmax=10,
    ymin=50, ymax=100,
    xtick={0,1,2,3,4,5,6,7,8,9,10},
    yticklabel={$\pgfmathprintnumber{\tick}\%$},
    legend pos=north west,
    legend cell align=left,
    grid=both,
    grid style={line width=.1pt, draw=gray!20},
    legend style={legend pos=south west, draw=none, fill=none, font=\fontsize{7}{5}\selectfont}
]

\addplot[smooth,mark=*,blue] 
    plot coordinates
    { (.5,60.9) (1,65.4) (3,76.1) (6,93.1) (10,95.5) };
    \addlegendentry{EESRGAN-FRCNN-End-to-End (SR-SR)}
    
\addplot[smooth,mark=*,red] 
    plot coordinates
    { (.5,56.7) (1,61.9) (3,71.9) (6,90.8) (10,93.6) };
    \addlegendentry{EESRGAN-FRCNN (SR-SR)}
    
    
\end{axis}
\end{tikzpicture}
\caption{COWC dataset}
\label{fig:dataset-COWC}
\end{subfigure}
\hfill
\hspace{15mm}
\centering
\begin{subfigure}[b]{.4\textwidth}
\centering
\centering
\begin{tikzpicture}
\begin{axis}[
    width=7.5cm,
    height=7cm,
    xlabel={Dataset Size ($\cdot10^2$)},
    ylabel={AP},
    xmin=0, xmax=14.91,
    ymin=50, ymax=100,
    xtick={0,2,4,6,8,10,12,14},
    yticklabel={$\pgfmathprintnumber{\tick}\%$},
    legend pos=north west,
    legend cell align=left,
    grid=both,
    grid style={line width=.1pt, draw=gray!20},
    legend style={legend pos=south west, draw=none, fill=none, font=\fontsize{7}{5}\selectfont}
]

\addplot[smooth,mark=*,blue] 
    plot coordinates
    { (1,68.2) (2,76.6) (4,78.2) (7.5,82) (14.91,83.2) };
    \addlegendentry{EESRGAN-FRCNN-End-to-End (SR-SR)}
    
\addplot[smooth,mark=*,red] 
    plot coordinates
    { (1,63.9) (2,74.1) (4,76.3) (7.5,80.5) (14.91,81.4) };
    \addlegendentry{EESRGAN-FRCNN (SR-SR)}
    
\end{axis}
\end{tikzpicture}
\caption{OGST dataset}
\label{fig:fig:dataset-OGT}
\end{subfigure}
\end{adjustbox}
\vspace{0.1cm}
\caption{AP (average precision) with varying number of training sets from the datasets. Plotted results show the detection performance of our proposed method (with and without end-to-end training) on SR (super-resolution) images.}
\label{fig:effect_dataset}
\end{figure}
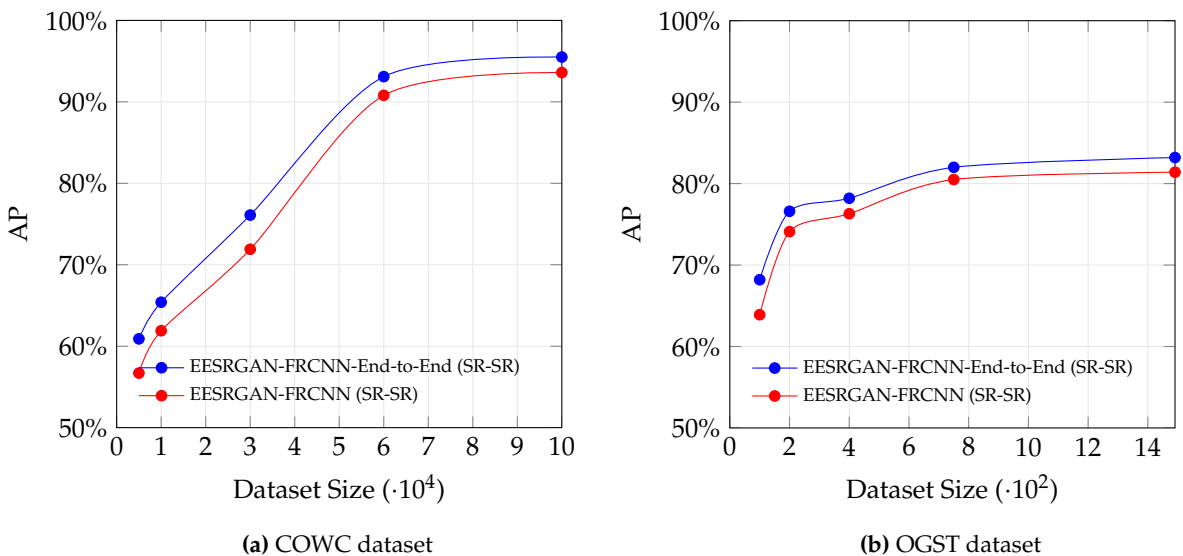

We trained our architecture with different training set sizes and tested with a fixed test set. In figure \ref{fig:effect_dataset}, we plot the AP values (IoU=0.5:0.95) against different numbers of labeled objects for both of our datasets (training data). We used five different dataset sizes: $\{500, 1000, 3000, 6000, 10000 (cars)\}$ and $\{100, 200, 400, 750, 1491 (tanks)\}$ to train our model with and without the end-to-end setting.  \par 
We got the highest AP value of 95.5\% with our full COWC training dataset (10000 cars), and we used the same test dataset (1000 cars) for all combinations of the training dataset (with end-to-end setting). We also used another set of 1000 labeled cars for validation. Using 6000 cars, we got an AP value near to the highest AP, as shown with the plot of AP versus dataset size (COWC). The AP value decreased significantly when we used only 3000 labeled cars as training data. We got the lowest AP using only 500 labeled cars, and the trend of AP was further decreasing as depicted in figure \ref{fig:effect_dataset}a. Therefore, we can infer that we needed around 6000 labeled cars to get precision higher than 90\% for the COWC dataset. We observed slightly lower AP values for all sizes of COWC datasets when we did not use the end-to-end setting, and we observed higher differences between the two settings (with and without end-to-end) when we used less than 6000 labeled cars. 

\captionsetup[figure]{width=.9\linewidth}

The OGST dataset gave 83.2\% AP (with end-to-end setting) using the full training dataset (1491 tanks), and we used 100 labeled tanks as test and same amount as validation data for all combinations of the training dataset. We got high AP values with 50\% of our full training dataset as depicted in \ref{fig:effect_dataset}b. AP values dropped below 80\% when we further decreased the training data. Similar to the COWC datasets, we also got comparatively lower AP values for all sizes of OGST datasets. We observed slightly higher differences between the two settings (with and without end-to-end) when the dataset consisted of less than 400 labeled tanks, as shown in the plot of AP versus dataset size (OGST dataset).
\par
We used 90\% of the OGST dataset for training while we used the 80\% of the COWC dataset for the same purpose. The accuracy of the testing data (OGST) slightly increased when we added more training data, as depicted in figure \ref{fig:effect_dataset}b. Therefore, we used a larger percentage of training data for the OGST dataset than for the COWC dataset, and it slightly helped to improve the relatively low accuracy of the OGST test data.

\subsubsection{Enhancement and Detection}

\begin{figure}[H]
\centering
\captionsetup{width=.9\linewidth}
\renewcommand{\arraystretch}{0.01}
\begin{tabular}{| @{} c @{} | @{} c @{} | @{} c @{}| @{} c @{}|}
\hline
\includegraphics[width = 3.72 cm, height = 3.2cm]{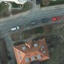} &
\includegraphics[width = 3.72 cm, height = 3.2cm]{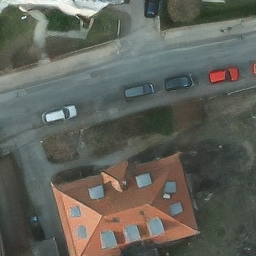} &
\includegraphics[width = 3.72 cm, height = 3.2cm]{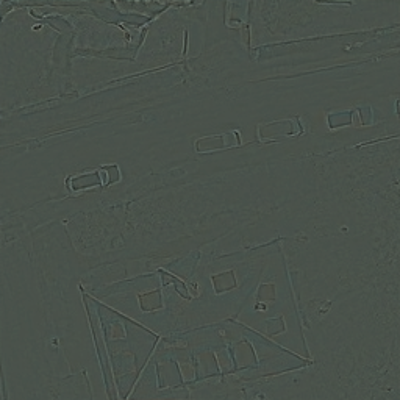} 
&
\includegraphics[width = 3.72 cm, height = 3.2cm]{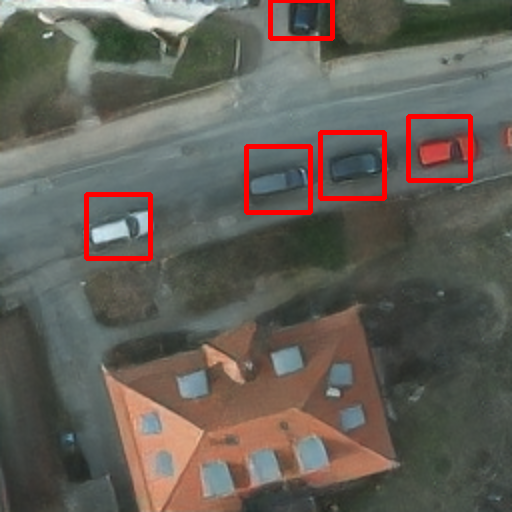} \\
\hline
\includegraphics[width = 3.72 cm, height = 3.2cm]{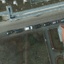} &
\includegraphics[width = 3.72 cm, height = 3.2cm]{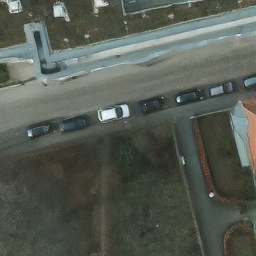} &
\includegraphics[width = 3.72 cm, height = 3.2cm]{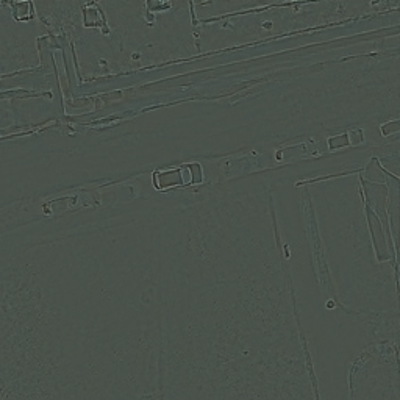}
&
\includegraphics[width = 3.72 cm, height = 3.2cm]{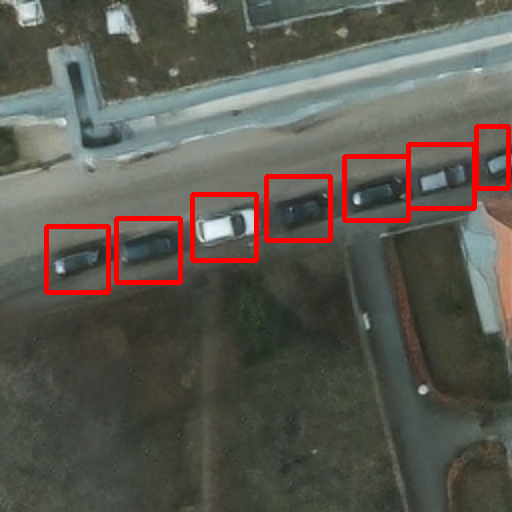}\\
\hline
\includegraphics[width = 3.72 cm, height = 3.2cm]{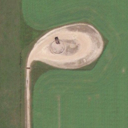} &
\includegraphics[width = 3.72 cm, height = 3.2cm]{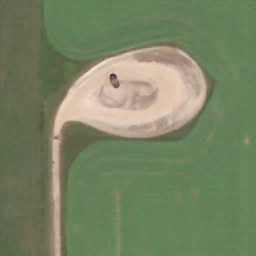} &
\includegraphics[width = 3.72 cm, height = 3.2cm]{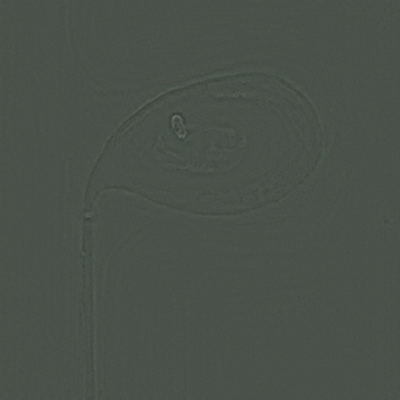} 
&
\includegraphics[width = 3.72 cm, height = 3.2cm]{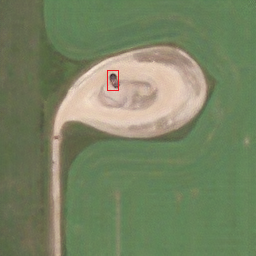}\\
\hline
\includegraphics[width = 3.72 cm, height = 3.2cm]{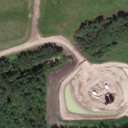} &
\includegraphics[width = 3.72 cm, height = 3.2cm]{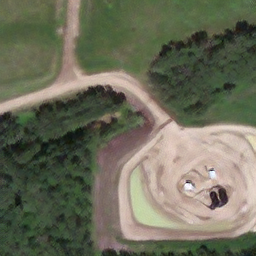} &
\includegraphics[width = 3.72 cm, height = 3.2cm]{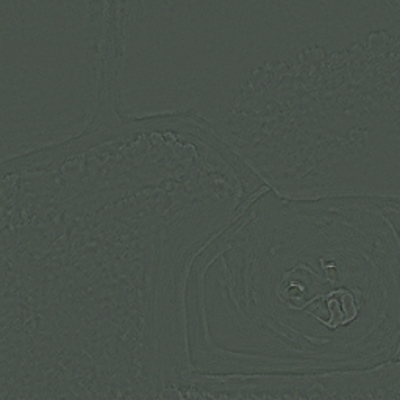} 
&
\includegraphics[width = 3.72 cm, height = 3.2cm]{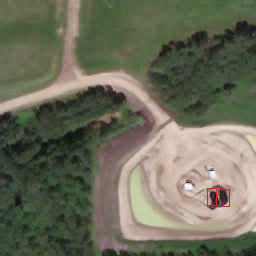}
\\
\hline
\arrayrulecolor{white}
\hline
\hline
\multicolumn{1}{c}{(a) Input LR image} & \multicolumn{1}{c}{(b) Generated SR image} & \multicolumn{1}{c}{(c) Enhanced edge} &
\multicolumn{1}{c}{(c) Detection}
\end{tabular}
\caption{Examples of SR (super-resolution) images that are generated from input LR (low-resolution) images are shown in (a) and (b). The enhanced edges and detection results are shown in (c) and (d).}
\label{fig:enhance_detect}
\end{figure}

In figure \ref{fig:enhance_detect}, we have shown our input LR images, corresponding generated SR image, enhanced edge information and final detection. The image enhancement has helped the detectors to get high AP values and also makes the images visually good enough to identify the objects easily. It is evident from the figure that the visual quality of the generated SR images is quite good compared to the corresponding LR images, and the FRCNN detector has detected most of the objects correctly. 

\subsubsection{Effects of Edge Consistency Loss ($\emph{L}_{edge\_cst}$)}
\begin{figure}[H]
\centering
\captionsetup{width=.9\linewidth}
\renewcommand{\arraystretch}{0.01}
\begin{tabular}{| @{} c @{} | @{} c @{} | @{} c @{}|}
\hline
\includegraphics[width = 1.5in]{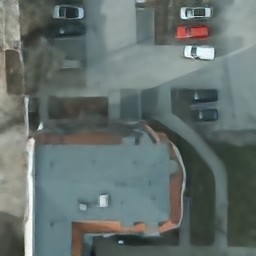} &
\includegraphics[width = 1.5 in]{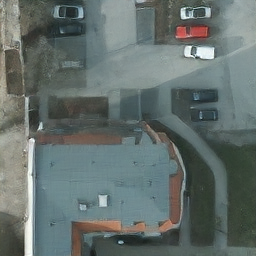} &
\includegraphics[width = 1.5in]{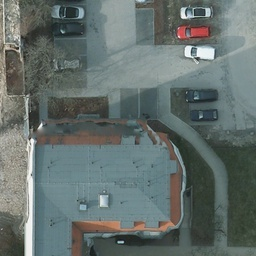} \\
\hline
\includegraphics[width = 1.5in]{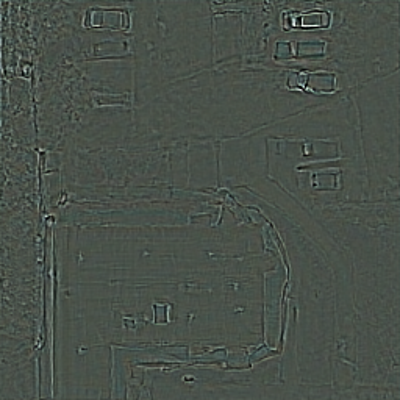} &
\includegraphics[width = 1.5 in]{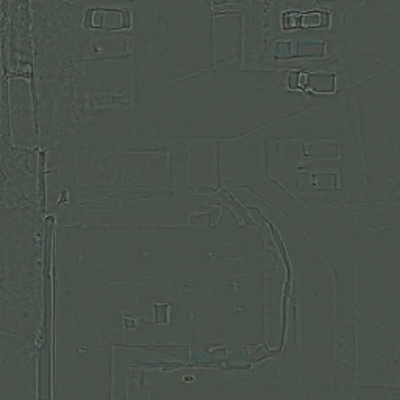} &
\includegraphics[width = 1.5in]{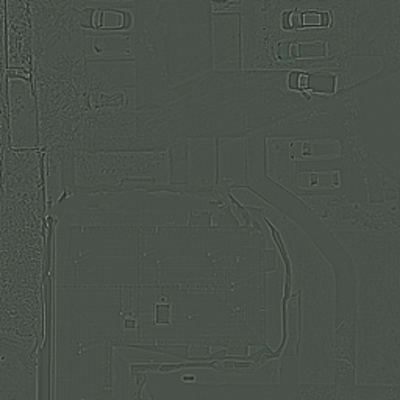} \\
\hline
\arrayrulecolor{white}
\hline
\hline
\multicolumn{1}{p{2.82cm}}{(a) Final SR image and enhanced edge with $\emph{L}_{img\_cst}$ loss} & 
\multicolumn{1}{p{2.82cm}}{(b) Final SR image and enhanced edge with $\emph{L}_{img\_cst}$ and $\emph{L}_{edge\_cst}$ losses} & 
\multicolumn{1}{p{2.82cm}}{(c) Ground truth HR image with extracted edge}
\end{tabular}
\caption{Effects of edge consistency loss ($\emph{L}_{edge\_cst}$) on final SR (super-resolution) images and enhanced edges compared to the extracted edges from HR (high-resolution) images.}
\label{fig:edge_loss}
\end{figure}

 In EEGAN \cite{EEGAN}, only image consistency loss ($\emph{L}_{img\_cst}$) was used for enhancing the edge information. This loss generated edge information with noise, and as a result, the final SR images became blurry. The blurry output with noisy edge using only $\emph{L}_{img\_cst}$ loss is shown in figure \ref{fig:edge_loss}a. The blurry final images gave lower detection accuracy compared to sharp outputs. 
 \par
 Therefore, we have introduced edge consistency loss ($\emph{L}_{edge\_cst}$) in addition to $\emph{L}_{img\_cst}$ loss that gives noise-free enhanced edge information similar to the edge extracted from ground truth images and the effects of the $\emph{L}_{edge\_cst}$ loss is shown in figure \ref{fig:edge_loss}b. The ground truth HR image with extracted edge is depicted in figure \ref{fig:edge_loss}c.

\section{Discussion}
\label{sec:discussion}
The detection results of our method presented in the previous section have indicated that our end-to-end SR-detector network improved detection accuracy compared to several other methods. Our method outperformed the standalone state-of-the-art methods such as SSD or faster R-CNN when implemented in low-resolution remote sensing imagery. We used EESRGAN, EEGAN, and ESRGAN as the SR network with the detectors. We showed that our EESRGAN with the detectors performed better than the other methods and the edge-enhancement helped to improve the detection accuracy. The AP improvement was higher in high IoUs and not so much in the lower IoUs. We have also showed that the precision increased with higher resolution. The improvement of AP values for the OGST dataset was lower than that for the COWC dataset because the area covered by a tank was slightly bigger than that of a car, and tanks sizes and colors were less diverse than the cars. 
\par
Our experimental results indicated that AP values of the output could be improved slightly with the increase of training data. The results also demonstrated that we could use less training data for both the datasets to get a similar level of accuracy that we obtained from our total training data. \par
The faster R-CNN detector gave us the best result, but it took a longer time than an SSD detector. If we need detection results from a vast area, then SSD would be the right choice sacrificing some amount of accuracy. 
\par
We had large numbers of cars from different regions in the COWC dataset, and we obtained high AP values using different IoUs. On the other hand, the OGST dataset needed more data to get a general detection result because we used data from a specific area and for a specific season and this was one of the limitations of our experiment. Most likely, more data from different regions and seasons would make our method more robust for the use-case of oil and gas storage tank detection. Another limitation of our experiment was that we showed the performance of the datasets that contain only one class with less variation. We would be looking forward to exploring the performance of our method on a broader range of object types and landscapes from different satellite datasets. \par

We have used LR-HR image pairs to train our architecture, and the LR images were generated artificially from the HR counterparts. To our knowledge, there is no suitable public satellite dataset that contains both real HR and real LR image pairs and ground truth bounding boxes for detecting small objects. Therefore, we have created the LR images which do not precisely correspond to true LR images. However, improvement of resolution through deep learning always improved object detection performance on remote sensing images (for both artificial and real low-resolution images), as discussed in the introduction and related works section of this paper \cite{small_object_enhance}. Impressive works \cite{cyclegan,real-low-resolution-image} exist in literature to create realistic LR images from HR images. For future work, we are looking forward to exploring the works to create more accurate LR images for training.


\section{Conclusions}
\label{sec:conclusions}
In this paper, we propose an end-to-end architecture that takes LR satellite imagery as input and gives object detection results as outputs. Our architecture contains a SR network and a detector network. We have used a different combination of SR systems and detectors to compare the AP values for detection using two different datasets. Our experimental results show that the proposed SR network with faster R-CNN has yielded the best results for small objects on satellite imagery. However, we need to add more diverse training data in the OGST dataset to make our model robust in detecting oil and gas storage tanks. We also need to explore diverse datasets and the techniques to create more realistic LR images. In conclusion, our method has combined different strategies to provide a better solution to the task of small-object detection on LR imagery.

\vspace{6pt} 



\authorcontributions{Conceptualization, J.R., N.R. and M.S.; methodology, J.R., N.R. and M.S.; software, J.R.; validation, J.R.; formal analysis, J.R.; investigation, J.R.; resources, N.R.; data curation, J.R., S.C. and D.C.; writing--original draft preparation, J.R.; writing--review and editing, J.R., N.R., M.S., S.C. and D.C.; visualization, J.R.; supervision, N.R. and M.S.; project administration, N.R.; funding acquisition, N.R., S.C. and D.C.}

\funding{This research was partially supported by Alberta Geological Survey (AGS) and NSERC discovery grant.}

\conflictsofinterest{The authors declare no conflict of interest.} 



\reftitle{References}


\externalbibliography{yes}



\begin{thebibliography}{-------}
\providecommand{\natexlab}[1]{#1}

\bibitem[Colomina and Molina(2014)]{military1}
Colomina, I.; Molina, P.
\newblock Unmanned aerial systems for photogrammetry and remote sensing: A
  review.
\newblock {\em ISPRS Journal of photogrammetry and remote sensing} {\bf 2014},
  {\em 92},~79--97.

\bibitem[Zhang \em{et~al.}(2016)Zhang, Du, Zhang, and Xu]{military2}
Zhang, F.; Du, B.; Zhang, L.; Xu, M.
\newblock Weakly supervised learning based on coupled convolutional neural
  networks for aircraft detection.
\newblock {\em IEEE Transactions on Geoscience and Remote Sensing} {\bf 2016},
  {\em 54},~5553--5563.

\bibitem[Fromm \em{et~al.}(2019)Fromm, Schubert, Castilla, Linke, and
  McDermid]{forestrySeedling}
Fromm, M.; Schubert, M.; Castilla, G.; Linke, J.; McDermid, G.
\newblock Automated Detection of Conifer Seedlings in Drone Imagery Using
  Convolutional Neural Networks.
\newblock {\em Remote Sensing} {\bf 2019}, {\em 11},~2585.

\bibitem[{Pang} \em{et~al.}(2019){Pang}, {Li}, {Shi}, {Xu}, and
  {Feng}]{TinyObjectFilter}
{Pang}, J.; {Li}, C.; {Shi}, J.; {Xu}, Z.; {Feng}, H.
\newblock $\mathcal{R}^2$ -CNN: Fast Tiny Object Detection in Large-Scale
  Remote Sensing Images.
\newblock {\em IEEE Transactions on Geoscience and Remote Sensing} {\bf 2019},
  {\em 57},~5512--5524.
\newblock
  doi:{\changeurlcolor{black}\href{https://doi.org/10.1109/TGRS.2019.2899955}{\detokenize{10.1109/TGRS.2019.2899955}}}.

\bibitem[Shermeyer and Van~Etten(2019)]{small_object_enhance}
Shermeyer, J.; Van~Etten, A.
\newblock The effects of super-resolution on object detection performance in
  satellite imagery.
\newblock  Proceedings of the IEEE Conference on Computer Vision and Pattern
  Recognition Workshops,  2019, pp. 0--10.

\bibitem[Russakovsky \em{et~al.}(2014)Russakovsky, Deng, Su, Krause, Satheesh,
  Ma, Huang, Karpathy, Khosla, Bernstein, Berg, and Fei-Fei]{imagenet}
Russakovsky, O.; Deng, J.; Su, H.; Krause, J.; Satheesh, S.; Ma, S.; Huang, Z.;
  Karpathy, A.; Khosla, A.; Bernstein, M.; Berg, A.C.; Fei-Fei, L.
\newblock ImageNet Large Scale Visual Recognition Challenge,  2014,
  \href{http://xxx.lanl.gov/abs/1409.0575}{{\normalfont
  [arXiv:cs.CV/1409.0575]}}.

\bibitem[Lin \em{et~al.}(2014)Lin, Maire, Belongie, Hays, Perona, Ramanan,
  Doll{\'a}r, and Zitnick]{mscoco}
Lin, T.Y.; Maire, M.; Belongie, S.; Hays, J.; Perona, P.; Ramanan, D.;
  Doll{\'a}r, P.; Zitnick, C.L.
\newblock Microsoft coco: Common objects in context.
\newblock  European conference on computer vision. Springer,  2014, pp.
  740--755.

\bibitem[Ren \em{et~al.}(2017)Ren, He, Girshick, and Sun]{FRCNN}
Ren, S.; He, K.; Girshick, R.; Sun, J.
\newblock Faster R-CNN: Towards Real-Time Object Detection with Region Proposal
  Networks.
\newblock {\em IEEE Transactions on Pattern Analysis and Machine Intelligence}
  {\bf 2017}, {\em 39},~1137–1149.
\newblock
  doi:{\changeurlcolor{black}\href{https://doi.org/10.1109/tpami.2016.2577031}{\detokenize{10.1109/tpami.2016.2577031}}}.

\bibitem[Lin \em{et~al.}(2017)Lin, Goyal, Girshick, He, and Dollar]{RetinaNet}
Lin, T.Y.; Goyal, P.; Girshick, R.; He, K.; Dollar, P.
\newblock Focal Loss for Dense Object Detection.
\newblock {\em 2017 IEEE International Conference on Computer Vision (ICCV)}
  {\bf 2017}.
\newblock
  doi:{\changeurlcolor{black}\href{https://doi.org/10.1109/iccv.2017.324}{\detokenize{10.1109/iccv.2017.324}}}.

\bibitem[Liu \em{et~al.}(2016)Liu, Anguelov, Erhan, Szegedy, Reed, Fu, and
  Berg]{SSD}
Liu, W.; Anguelov, D.; Erhan, D.; Szegedy, C.; Reed, S.; Fu, C.Y.; Berg, A.C.
\newblock SSD: Single Shot MultiBox Detector.
\newblock {\em Lecture Notes in Computer Science} {\bf 2016}, p. 21–37.
\newblock
  doi:{\changeurlcolor{black}\href{https://doi.org/10.1007/978-3-319-46448-0_2}{\detokenize{10.1007/978-3-319-46448-0_2}}}.

\bibitem[Redmon \em{et~al.}(2016)Redmon, Divvala, Girshick, and Farhadi]{yolo}
Redmon, J.; Divvala, S.; Girshick, R.; Farhadi, A.
\newblock You Only Look Once: Unified, Real-Time Object Detection.
\newblock {\em 2016 IEEE Conference on Computer Vision and Pattern Recognition
  (CVPR)} {\bf 2016}.
\newblock
  doi:{\changeurlcolor{black}\href{https://doi.org/10.1109/cvpr.2016.91}{\detokenize{10.1109/cvpr.2016.91}}}.

\bibitem[{Ji} \em{et~al.}(2019){Ji}, {Gao}, {Mei}, and
  {Ramesh}]{vehicleSuperResolution}
{Ji}, H.; {Gao}, Z.; {Mei}, T.; {Ramesh}, B.
\newblock Vehicle Detection in Remote Sensing Images Leveraging on Simultaneous
  Super-Resolution.
\newblock {\em IEEE Geoscience and Remote Sensing Letters} {\bf 2019}, pp.
  1--5.
\newblock
  doi:{\changeurlcolor{black}\href{https://doi.org/10.1109/LGRS.2019.2930308}{\detokenize{10.1109/LGRS.2019.2930308}}}.

\bibitem[Tayara \em{et~al.}(2017)Tayara, Soo, and Chong]{Vehicle1}
Tayara, H.; Soo, K.G.; Chong, K.T.
\newblock Vehicle detection and counting in high-resolution aerial images using
  convolutional regression neural network.
\newblock {\em IEEE Access} {\bf 2017}, {\em 6},~2220--2230.

\bibitem[Yu and Shi(2015)]{vehicle2}
Yu, X.; Shi, Z.
\newblock Vehicle detection in remote sensing imagery based on salient
  information and local shape feature.
\newblock {\em Optik-International Journal for Light and Electron Optics} {\bf
  2015}, {\em 126},~2485--2490.

\bibitem[Stankov and He(2014)]{building}
Stankov, K.; He, D.C.
\newblock Detection of buildings in multispectral very high spatial resolution
  images using the percentage occupancy hit-or-miss transform.
\newblock {\em IEEE Journal of Selected Topics in Applied Earth Observations
  and Remote Sensing} {\bf 2014}, {\em 7},~4069--4080.

\bibitem[Ok and Ba{\c{s}}eski(2015)]{storage_tank}
Ok, A.O.; Ba{\c{s}}eski, E.
\newblock Circular oil tank detection from panchromatic satellite images: A new
  automated approach.
\newblock {\em IEEE Geoscience and Remote Sensing Letters} {\bf 2015}, {\em
  12},~1347--1351.

\bibitem[Dong \em{et~al.}(2016)Dong, Loy, He, and Tang]{SRCNN}
Dong, C.; Loy, C.C.; He, K.; Tang, X.
\newblock Image Super-Resolution Using Deep Convolutional Networks.
\newblock {\em IEEE Transactions on Pattern Analysis and Machine Intelligence}
  {\bf 2016}, {\em 38},~295–307.
\newblock
  doi:{\changeurlcolor{black}\href{https://doi.org/10.1109/tpami.2015.2439281}{\detokenize{10.1109/tpami.2015.2439281}}}.

\bibitem[Kim \em{et~al.}(2016)Kim, Lee, and Lee]{VDSR}
Kim, J.; Lee, J.K.; Lee, K.M.
\newblock Accurate Image Super-Resolution Using Very Deep Convolutional
  Networks.
\newblock {\em 2016 IEEE Conference on Computer Vision and Pattern Recognition
  (CVPR)} {\bf 2016}.
\newblock
  doi:{\changeurlcolor{black}\href{https://doi.org/10.1109/cvpr.2016.182}{\detokenize{10.1109/cvpr.2016.182}}}.

\bibitem[Goodfellow \em{et~al.}(2014)Goodfellow, Pouget-Abadie, Mirza, Xu,
  Warde-Farley, Ozair, Courville, and Bengio]{GAN}
Goodfellow, I.; Pouget-Abadie, J.; Mirza, M.; Xu, B.; Warde-Farley, D.; Ozair,
  S.; Courville, A.; Bengio, Y.
\newblock Generative adversarial nets.
\newblock  Advances in neural information processing systems,  2014, pp.
  2672--2680.

\bibitem[Ledig \em{et~al.}(2017)Ledig, Theis, Huszar, Caballero, Cunningham,
  Acosta, Aitken, Tejani, Totz, Wang, and et~al.]{SRGAN}
Ledig, C.; Theis, L.; Huszar, F.; Caballero, J.; Cunningham, A.; Acosta, A.;
  Aitken, A.; Tejani, A.; Totz, J.; Wang, Z.; et~al..
\newblock Photo-Realistic Single Image Super-Resolution Using a Generative
  Adversarial Network.
\newblock {\em 2017 IEEE Conference on Computer Vision and Pattern Recognition
  (CVPR)} {\bf 2017}.
\newblock
  doi:{\changeurlcolor{black}\href{https://doi.org/10.1109/cvpr.2017.19}{\detokenize{10.1109/cvpr.2017.19}}}.

\bibitem[Wang \em{et~al.}(2019)Wang, Yu, Wu, Gu, Liu, Dong, Qiao, and
  Loy]{ESRGAN}
Wang, X.; Yu, K.; Wu, S.; Gu, J.; Liu, Y.; Dong, C.; Qiao, Y.; Loy, C.C.
\newblock ESRGAN: Enhanced Super-Resolution Generative Adversarial Networks.
\newblock {\em Computer Vision – ECCV 2018 Workshops} {\bf 2019}, p. 63–79.
\newblock
  doi:{\changeurlcolor{black}\href{https://doi.org/10.1007/978-3-030-11021-5_5}{\detokenize{10.1007/978-3-030-11021-5_5}}}.

\bibitem[{Jiang} \em{et~al.}(2019{\natexlab{a}}){Jiang}, {Wang}, {Yi}, {Wang},
  {Lu}, and {Jiang}]{EEGAN}
{Jiang}, K.; {Wang}, Z.; {Yi}, P.; {Wang}, G.; {Lu}, T.; {Jiang}, J.
\newblock Edge-Enhanced GAN for Remote Sensing Image Superresolution.
\newblock {\em IEEE Transactions on Geoscience and Remote Sensing} {\bf 2019},
  {\em 57},~5799--5812.
\newblock
  doi:{\changeurlcolor{black}\href{https://doi.org/10.1109/TGRS.2019.2902431}{\detokenize{10.1109/TGRS.2019.2902431}}}.

\bibitem[{Jiang} \em{et~al.}(2019{\natexlab{b}}){Jiang}, {Ma}, {Wang}, {Chen},
  and {Liu}]{edge_land_cover1}
{Jiang}, J.; {Ma}, J.; {Wang}, Z.; {Chen}, C.; {Liu}, X.
\newblock Hyperspectral Image Classification in the Presence of Noisy Labels.
\newblock {\em IEEE Transactions on Geoscience and Remote Sensing} {\bf 2019},
  {\em 57},~851--865.
\newblock
  doi:{\changeurlcolor{black}\href{https://doi.org/10.1109/TGRS.2018.2861992}{\detokenize{10.1109/TGRS.2018.2861992}}}.

\bibitem[Tong \em{et~al.}(2017)Tong, Tong, Jiang, and Zhang]{edge_land_cover2}
Tong, F.; Tong, H.; Jiang, J.; Zhang, Y.
\newblock Multiscale union regions adaptive sparse representation for
  hyperspectral image classification.
\newblock {\em Remote Sensing} {\bf 2017}, {\em 9},~872.

\bibitem[Zhan \em{et~al.}(2007)Zhan, Duan, Xu, Song, and Luo]{edge_important}
Zhan, C.; Duan, X.; Xu, S.; Song, Z.; Luo, M.
\newblock An improved moving object detection algorithm based on frame
  difference and edge detection.
\newblock  Fourth International Conference on Image and Graphics (ICIG 2007).
  IEEE,  2007, pp. 519--523.

\bibitem[Mao \em{et~al.}(2018)Mao, Wang, Wang, Zhang, and Ma]{edge_CNN1}
Mao, Q.; Wang, S.; Wang, S.; Zhang, X.; Ma, S.
\newblock Enhanced image decoding via edge-preserving generative adversarial
  networks.
\newblock  2018 IEEE International Conference on Multimedia and Expo (ICME).
  IEEE,  2018, pp. 1--6.

\bibitem[Yang \em{et~al.}(2017)Yang, Feng, Yang, Zhao, Liu, Guo, and
  Yan]{edge_CNN2}
Yang, W.; Feng, J.; Yang, J.; Zhao, F.; Liu, J.; Guo, Z.; Yan, S.
\newblock Deep Edge Guided Recurrent Residual Learning for Image
  Super-Resolution.
\newblock {\em IEEE Transactions on Image Processing} {\bf 2017}, {\em
  26},~5895–5907.
\newblock
  doi:{\changeurlcolor{black}\href{https://doi.org/10.1109/tip.2017.2750403}{\detokenize{10.1109/tip.2017.2750403}}}.

\bibitem[{Kamgar-Parsi} \em{et~al.}(1999){Kamgar-Parsi}, {Kamgar-Parsi}, and
  {Rosenfeld}]{laplacian_operator}
{Kamgar-Parsi}, B.; {Kamgar-Parsi}, B.; {Rosenfeld}, A.
\newblock Optimally isotropic Laplacian operator.
\newblock {\em IEEE Transactions on Image Processing} {\bf 1999}, {\em
  8},~1467--1472.
\newblock
  doi:{\changeurlcolor{black}\href{https://doi.org/10.1109/83.791975}{\detokenize{10.1109/83.791975}}}.

\bibitem[lan()]{landsat-8}
Landsat 8.
\newblock \url{https://www.usgs.gov/land-resources/nli/landsat/landsat-8}.
\newblock Accessed: 2020-02-11.

\bibitem[sen()]{sentinel-2}
Sentinel-2.
\newblock
  \url{http://www.esa.int/Applications/Observing_the_Earth/Copernicus/Sentinel-2}.
\newblock Accessed: 2020-02-11.

\bibitem[Mundhenk \em{et~al.}(2016)Mundhenk, Konjevod, Sakla, and Boakye]{cowc}
Mundhenk, T.N.; Konjevod, G.; Sakla, W.A.; Boakye, K.
\newblock A large contextual dataset for classification, detection and counting
  of cars with deep learning.
\newblock  European Conference on Computer Vision. Springer,  2016, pp.
  785--800.

\bibitem[Rabbi \em{et~al.}(2020)Rabbi, Chowdhury, and
  Chao]{oil-gas-tank-dataset}
Rabbi, J.; Chowdhury, S.; Chao, D.
\newblock Oil and Gas Tank Dataset.
\newblock Mendeley Data, V3,  2020.
\newblock
  doi:{\changeurlcolor{black}\href{https://doi.org/10.17632/bkxj8z84m9.3}{\detokenize{10.17632/bkxj8z84m9.3}}}.

\bibitem[Jolicoeur-Martineau(2018)]{relativisticGAN}
Jolicoeur-Martineau, A.
\newblock The relativistic discriminator: a key element missing from standard
  GAN,  2018,  \href{http://xxx.lanl.gov/abs/1807.00734}{{\normalfont
  [arXiv:cs.LG/1807.00734]}}.

\bibitem[Charbonnier \em{et~al.}(1994)Charbonnier, Blanc-F{\'e}raud, Aubert,
  and Barlaud]{Charbonnier1994TwoDH}
Charbonnier, P.; Blanc-F{\'e}raud, L.; Aubert, G.; Barlaud, M.
\newblock Two deterministic half-quadratic regularization algorithms for
  computed imaging.
\newblock {\em Proceedings of 1st International Conference on Image Processing}
  {\bf 1994}, {\em 2},~168--172 vol.2.

\bibitem[AER()]{AER}
Alberta Energy Regulator.
\newblock \url{https://www.aer.ca}.
\newblock Accessed: 2020-02-05.

\bibitem[Tai \em{et~al.}(2017)Tai, Yang, Liu, and Xu]{dense_network}
Tai, Y.; Yang, J.; Liu, X.; Xu, C.
\newblock MemNet: A Persistent Memory Network for Image Restoration.
\newblock {\em 2017 IEEE International Conference on Computer Vision (ICCV)}
  {\bf 2017}.
\newblock
  doi:{\changeurlcolor{black}\href{https://doi.org/10.1109/iccv.2017.486}{\detokenize{10.1109/iccv.2017.486}}}.

\bibitem[Zhang \em{et~al.}(2018)Zhang, Tian, Kong, Zhong, and
  Fu]{residual_dense_network}
Zhang, Y.; Tian, Y.; Kong, Y.; Zhong, B.; Fu, Y.
\newblock Residual Dense Network for Image Super-Resolution.
\newblock {\em 2018 IEEE/CVF Conference on Computer Vision and Pattern
  Recognition} {\bf 2018}.
\newblock
  doi:{\changeurlcolor{black}\href{https://doi.org/10.1109/cvpr.2018.00262}{\detokenize{10.1109/cvpr.2018.00262}}}.

\bibitem[He \em{et~al.}(2015)He, Zhang, Ren, and Sun]{He_2015}
He, K.; Zhang, X.; Ren, S.; Sun, J.
\newblock Delving Deep into Rectifiers: Surpassing Human-Level Performance on
  ImageNet Classification.
\newblock {\em 2015 IEEE International Conference on Computer Vision (ICCV)}
  {\bf 2015}.
\newblock
  doi:{\changeurlcolor{black}\href{https://doi.org/10.1109/iccv.2015.123}{\detokenize{10.1109/iccv.2015.123}}}.

\bibitem[Lim \em{et~al.}(2017)Lim, Son, Kim, Nah, and Lee]{EDSR}
Lim, B.; Son, S.; Kim, H.; Nah, S.; Lee, K.M.
\newblock Enhanced Deep Residual Networks for Single Image Super-Resolution.
\newblock {\em 2017 IEEE Conference on Computer Vision and Pattern Recognition
  Workshops (CVPRW)} {\bf 2017}.
\newblock
  doi:{\changeurlcolor{black}\href{https://doi.org/10.1109/cvprw.2017.151}{\detokenize{10.1109/cvprw.2017.151}}}.

\bibitem[Liebel and K{\"o}rner(2016)]{super-resolution-multispectral}
Liebel, L.; K{\"o}rner, M.
\newblock Single-image super resolution for multispectral remote sensing data
  using convolutional neural networks.
\newblock {\em ISPRS-International Archives of the Photogrammetry, Remote
  Sensing and Spatial Information Sciences} {\bf 2016}, {\em 41},~883--890.

\bibitem[Tayara and Chong(2018)]{detectors_two_type}
Tayara, H.; Chong, K.
\newblock Object detection in very high-resolution aerial images using
  one-stage densely connected feature pyramid network.
\newblock {\em Sensors} {\bf 2018}, {\em 18},~3341.

\bibitem[Girshick \em{et~al.}(2014)Girshick, Donahue, Darrell, and Malik]{RCNN}
Girshick, R.; Donahue, J.; Darrell, T.; Malik, J.
\newblock Rich Feature Hierarchies for Accurate Object Detection and Semantic
  Segmentation.
\newblock {\em 2014 IEEE Conference on Computer Vision and Pattern Recognition}
  {\bf 2014}.
\newblock
  doi:{\changeurlcolor{black}\href{https://doi.org/10.1109/cvpr.2014.81}{\detokenize{10.1109/cvpr.2014.81}}}.

\bibitem[Girshick(2015)]{Fast_RCNN}
Girshick, R.
\newblock Fast R-CNN.
\newblock {\em 2015 IEEE International Conference on Computer Vision (ICCV)}
  {\bf 2015}.
\newblock
  doi:{\changeurlcolor{black}\href{https://doi.org/10.1109/iccv.2015.169}{\detokenize{10.1109/iccv.2015.169}}}.

\bibitem[Li \em{et~al.}(2019)Li, Mou, Xu, Zhang, and Zhu]{vehicle_related_work}
Li, Q.; Mou, L.; Xu, Q.; Zhang, Y.; Zhu, X.X.
\newblock R3-Net: A Deep Network for Multioriented Vehicle Detection in Aerial
  Images and Videos.
\newblock {\em IEEE Transactions on Geoscience and Remote Sensing} {\bf 2019},
  {\em 57},~5028–5042.
\newblock
  doi:{\changeurlcolor{black}\href{https://doi.org/10.1109/tgrs.2019.2895362}{\detokenize{10.1109/tgrs.2019.2895362}}}.

\bibitem[Ammour \em{et~al.}(2017)Ammour, Alhichri, Bazi, Benjdira, Alajlan, and
  Zuair]{vehicle_related_work1}
Ammour, N.; Alhichri, H.; Bazi, Y.; Benjdira, B.; Alajlan, N.; Zuair, M.
\newblock Deep learning approach for car detection in UAV imagery.
\newblock {\em Remote Sensing} {\bf 2017}, {\em 9},~312.

\bibitem[Ren \em{et~al.}(2018)Ren, Zhu, and Xiao]{small-object}
Ren, Y.; Zhu, C.; Xiao, S.
\newblock Small object detection in optical remote sensing images via modified
  faster R-CNN.
\newblock {\em Applied Sciences} {\bf 2018}, {\em 8},~813.

\bibitem[Tang \em{et~al.}(2017)Tang, Zhou, Deng, Zou, and
  Lei]{vehicle-hard-negative}
Tang, T.; Zhou, S.; Deng, Z.; Zou, H.; Lei, L.
\newblock Vehicle detection in aerial images based on region convolutional
  neural networks and hard negative example mining.
\newblock {\em Sensors} {\bf 2017}, {\em 17},~336.

\bibitem[Chen \em{et~al.}(2018)Chen, Zhang, and Ouyang]{plane-ssd}
Chen, Z.; Zhang, T.; Ouyang, C.
\newblock End-to-end airplane detection using transfer learning in remote
  sensing images.
\newblock {\em Remote Sensing} {\bf 2018}, {\em 10},~139.

\bibitem[Radovic \em{et~al.}(2017)Radovic, Adarkwa, and
  Wang]{yolo-parameter-tuning}
Radovic, M.; Adarkwa, O.; Wang, Q.
\newblock Object recognition in aerial images using convolutional neural
  networks.
\newblock {\em Journal of Imaging} {\bf 2017}, {\em 3},~21.

\bibitem[Li \em{et~al.}(2017)Li, Fu, Yu, and Cracknell]{palm-tree-cnn}
Li, W.; Fu, H.; Yu, L.; Cracknell, A.
\newblock Deep learning based oil palm tree detection and counting for
  high-resolution remote sensing images.
\newblock {\em Remote Sensing} {\bf 2017}, {\em 9},~22.

\bibitem[Lin \em{et~al.}(2017)Lin, Dollar, Girshick, He, Hariharan, and
  Belongie]{FPNresnet50}
Lin, T.Y.; Dollar, P.; Girshick, R.; He, K.; Hariharan, B.; Belongie, S.
\newblock Feature Pyramid Networks for Object Detection.
\newblock {\em 2017 IEEE Conference on Computer Vision and Pattern Recognition
  (CVPR)} {\bf 2017}.
\newblock
  doi:{\changeurlcolor{black}\href{https://doi.org/10.1109/cvpr.2017.106}{\detokenize{10.1109/cvpr.2017.106}}}.

\bibitem[Liu \em{et~al.}(2018)Liu, Huang, et~al.]{RFBNet}
Liu, S.; Huang, D.; others.
\newblock Receptive field block net for accurate and fast object detection.
\newblock  Proceedings of the European Conference on Computer Vision (ECCV),
  2018, pp. 385--400.

\bibitem[Zhang \em{et~al.}(2018)Zhang, Wen, Bian, Lei, and Li]{refineDet}
Zhang, S.; Wen, L.; Bian, X.; Lei, Z.; Li, S.Z.
\newblock Single-shot refinement neural network for object detection.
\newblock  Proceedings of the IEEE conference on computer vision and pattern
  recognition,  2018, pp. 4203--4212.

\bibitem[Li and Zhou(2017)]{fssd}
Li, Z.; Zhou, F.
\newblock FSSD: feature fusion single shot multibox detector.
\newblock {\em arXiv preprint arXiv:1712.00960} {\bf 2017}.

\bibitem[Zhu \em{et~al.}(2019)Zhu, Zhang, Wang, Wen, Shi, Bo, and
  Mei]{scratchdet}
Zhu, R.; Zhang, S.; Wang, X.; Wen, L.; Shi, H.; Bo, L.; Mei, T.
\newblock ScratchDet: Training single-shot object detectors from scratch.
\newblock  Proceedings of the IEEE conference on computer vision and pattern
  recognition,  2019, pp. 2268--2277.

\bibitem[Yang \em{et~al.}(2018)Yang, Sun, Fu, Yang, Sun, Yan, and
  Guo]{small-ship}
Yang, X.; Sun, H.; Fu, K.; Yang, J.; Sun, X.; Yan, M.; Guo, Z.
\newblock Automatic ship detection in remote sensing images from google earth
  of complex scenes based on multiscale rotation dense feature pyramid
  networks.
\newblock {\em Remote Sensing} {\bf 2018}, {\em 10},~132.

\bibitem[Zhao \em{et~al.}(2019)Zhao, Zheng, Xu, and Wu]{object-review}
Zhao, Z.Q.; Zheng, P.; Xu, S.t.; Wu, X.
\newblock Object detection with deep learning: A review.
\newblock {\em IEEE transactions on neural networks and learning systems} {\bf
  2019}, {\em 30},~3212--3232.

\bibitem[Li \em{et~al.}(2020)Li, Wan, Cheng, Meng, and
  Han]{object-review-remote}
Li, K.; Wan, G.; Cheng, G.; Meng, L.; Han, J.
\newblock Object detection in optical remote sensing images: A survey and a new
  benchmark.
\newblock {\em ISPRS Journal of Photogrammetry and Remote Sensing} {\bf 2020},
  {\em 159},~296--307.

\bibitem[Bai \em{et~al.}(2018)Bai, Zhang, Ding, and Ghanem]{enhance-detect1}
Bai, Y.; Zhang, Y.; Ding, M.; Ghanem, B.
\newblock Sod-mtgan: Small object detection via multi-task generative
  adversarial network.
\newblock  Proceedings of the European Conference on Computer Vision (ECCV),
  2018, pp. 206--221.

\bibitem[Haris \em{et~al.}(2018)Haris, Shakhnarovich, and
  Ukita]{enhance-detect2}
Haris, M.; Shakhnarovich, G.; Ukita, N.
\newblock Task-driven super resolution: Object detection in low-resolution
  images.
\newblock {\em arXiv preprint arXiv:1803.11316} {\bf 2018}.

\bibitem[Zhu \em{et~al.}(2017)Zhu, Park, Isola, and Efros]{cyclegan}
Zhu, J.Y.; Park, T.; Isola, P.; Efros, A.A.
\newblock Unpaired image-to-image translation using cycle-consistent
  adversarial networks.
\newblock  Proceedings of the IEEE international conference on computer vision,
   2017, pp. 2223--2232.

\bibitem[Simonyan and Zisserman(2014)]{vgg19}
Simonyan, K.; Zisserman, A.
\newblock Very Deep Convolutional Networks for Large-Scale Image Recognition,
  2014,  \href{http://xxx.lanl.gov/abs/1409.1556}{{\normalfont
  [arXiv:cs.CV/1409.1556]}}.

\bibitem[He \em{et~al.}(2015)He, Zhang, Ren, and Sun]{PRELU}
He, K.; Zhang, X.; Ren, S.; Sun, J.
\newblock Delving Deep into Rectifiers: Surpassing Human-Level Performance on
  ImageNet Classification.
\newblock {\em 2015 IEEE International Conference on Computer Vision (ICCV)}
  {\bf 2015}.
\newblock
  doi:{\changeurlcolor{black}\href{https://doi.org/10.1109/iccv.2015.123}{\detokenize{10.1109/iccv.2015.123}}}.

\bibitem[Lai \em{et~al.}(2017)Lai, Huang, Ahuja, and Yang]{charbonnier_penalty}
Lai, W.S.; Huang, J.B.; Ahuja, N.; Yang, M.H.
\newblock Deep Laplacian Pyramid Networks for Fast and Accurate
  Super-Resolution.
\newblock {\em 2017 IEEE Conference on Computer Vision and Pattern Recognition
  (CVPR)} {\bf 2017}.
\newblock
  doi:{\changeurlcolor{black}\href{https://doi.org/10.1109/cvpr.2017.618}{\detokenize{10.1109/cvpr.2017.618}}}.

\bibitem[Kingma and Ba(2014)]{adam}
Kingma, D.P.; Ba, J.
\newblock Adam: A method for stochastic optimization.
\newblock {\em arXiv preprint arXiv:1412.6980} {\bf 2014}.

\bibitem[Paszke \em{et~al.}(2019)Paszke, Gross, Massa, Lerer, Bradbury, Chanan,
  Killeen, Lin, Gimelshein, Antiga, Desmaison, Kopf, Yang, DeVito, Raison,
  Tejani, Chilamkurthy, Steiner, Fang, Bai, and Chintala]{pytorch}
Paszke, A.; Gross, S.; Massa, F.; Lerer, A.; Bradbury, J.; Chanan, G.; Killeen,
  T.; Lin, Z.; Gimelshein, N.; Antiga, L.; Desmaison, A.; Kopf, A.; Yang, E.;
  DeVito, Z.; Raison, M.; Tejani, A.; Chilamkurthy, S.; Steiner, B.; Fang, L.;
  Bai, J.; Chintala, S.
\newblock PyTorch: An Imperative Style, High-Performance Deep Learning Library.
  In {\em Advances in Neural Information Processing Systems 32}; Wallach, H.;
  Larochelle, H.; Beygelzimer, A.; d'Alch\'{e} Buc, F.; Fox, E.; Garnett, R.,
  Eds.; Curran Associates, Inc.,  2019; pp. 8024--8035.

\bibitem[Rabbi(2020)]{Filter_Enhance_Detect}
Rabbi, J.
\newblock {Edge Enhanced GAN with Faster RCNN for end-to-end object detection
  from remote sensing imagery}.
\newblock \url{https://github.com/Jakaria08/Filter_Enhance_Detect},  2020.

\bibitem[AGS()]{AGS}
Alberta Geological Survey.
\newblock \url{https://ags.aer.ca}.
\newblock Accessed: 2020-02-05.

\bibitem[Chowdhury \em{et~al.}(2017)Chowdhury, Chao, Shipman, and
  Wulder]{ags_subir}
Chowdhury, S.; Chao, D.K.; Shipman, T.C.; Wulder, M.A.
\newblock Utilization of Landsat data to quantify land-use and land-cover
  changes related to oil and gas activities in West-Central Alberta from 2005
  to 2013.
\newblock {\em GIScience \& Remote Sensing} {\bf 2017}, {\em 54},~700--720.

\bibitem[bin()]{bing}
Bing Map.
\newblock \url{https://www.bing.com/maps}.
\newblock Accessed: 2020-02-05.

\bibitem[Bulat \em{et~al.}(2018)Bulat, Yang, and
  Tzimiropoulos]{real-low-resolution-image}
Bulat, A.; Yang, J.; Tzimiropoulos, G.
\newblock To learn image super-resolution, use a gan to learn how to do image
  degradation first.
\newblock  Proceedings of the European conference on computer vision (ECCV),
  2018, pp. 185--200.

\end{thebibliography}
\end{document}